\documentclass[10pt,twocolumn,letterpaper]{article}

\usepackage[pagenumbers]{cvpr} 

\newif\ifnotanon
\notanontrue

\newif\ifarxiv
\arxivtrue

\newcommand{\xxnote}[3]{}
\ifx\hidenotes\undefined
  \renewcommand{\xxnote}[3]{\color{#2}{#1: #3}}
\fi

\newcommand{\methodname}{\textbf{\texttt{RADSeg}}}

\usepackage[accsupp]{axessibility}  

\definecolor{cvprblue}{rgb}{0.21,0.49,0.74}
\usepackage[pagebackref,breaklinks,colorlinks,allcolors=cvprblue]{hyperref}
\usepackage{multirow}
\usepackage[dvipsnames]{xcolor}
\usepackage{xcolor,colortbl}

\def\paperID{XXXXX} 
\def\confName{CVPR}
\def\confYear{2026}
\newcommand{\webpage}{radseg-ovss.github.io}

\definecolor{navy}{RGB}{0,119,182}
\definecolor{ourpurple}{RGB}{92,66,154}
\ifarxiv
\title{RADSeg: Unleashing Parameter and Compute Efficient Zero-Shot Open-Vocabulary Segmentation Using Agglomerative Models \\ [2pt] \large{\href{https://\webpage/}{\color{ourpurple}{\webpage}}}}
\else
\title{RADSeg: Unleashing Parameter and Compute Efficient Zero-Shot Open-Vocabulary Segmentation Using Agglomerative Models}
\fi

\ifarxiv
\newcommand{\authorhref}[3][navy]{\href{#2}{\color{#1}{#3}}}
\vspace{-2em}
\author{
\authorhref{https://oasisartisan.github.io/}{Omar Alama}\textsuperscript{*1},
\authorhref{https://www.linkedin.com/in/darshil-jariwala/}{Darshil Jariwala}\textsuperscript{*2},
\authorhref{https://www.linkedin.com/in/avigyan-bhattacharya}{Avigyan Bhattacharya}\textsuperscript{*1}, \\
\authorhref{https://seungchan-kim.github.io/}{Seungchan Kim}\textsuperscript{1}, 
\authorhref{https://theairlab.org/team/wenshan/}{Wenshan Wang}\textsuperscript{1}, 
\authorhref{https://theairlab.org/team/sebastian/}{Sebastian Scherer}\textsuperscript{1}
\\[5 pt]
\textsuperscript{1}\href{https://www.ri.cmu.edu/}{\color{ourpurple}{Carnegie Mellon University}}
\textsuperscript{2}\href{https://www.iiit.ac.in/}{\color{ourpurple}{IIIT Hyderabad}}\thanks{Equal contribution}
}
\else
\author{
Omar Alama$^{*1}$ \quad Darshil Jariwala$^{*2}$ \quad Avigyan Bhattacharya$^{*1}$ \\
Seungchan Kim$^{1}$ \quad Wenshan Wang$^{1}$ \quad Sebastian Scherer$^{1}$ \\[4pt]
$^1$ Carnegie Mellon University \quad $^2$ IIIT Hyderabad
\thanks{Equal contribution}
}
\fi

\begin{document}

\ifarxiv
\makeatletter
\renewcommand{\@maketitle}{%
  \newpage
  \null
  \vskip -1.5em   
  \begin{center}%
  \let \footnote \thanks
    {\LARGE \@title \par}%
    \vskip 0.6em   
    {\large \@author}%
  \end{center}%
  \par
  \vskip 1em
  \vspace{-5mm}
\captionsetup{type=figure, singlelinecheck=false}
\begin{tabular}{cccc}
\includegraphics[width=\textwidth]{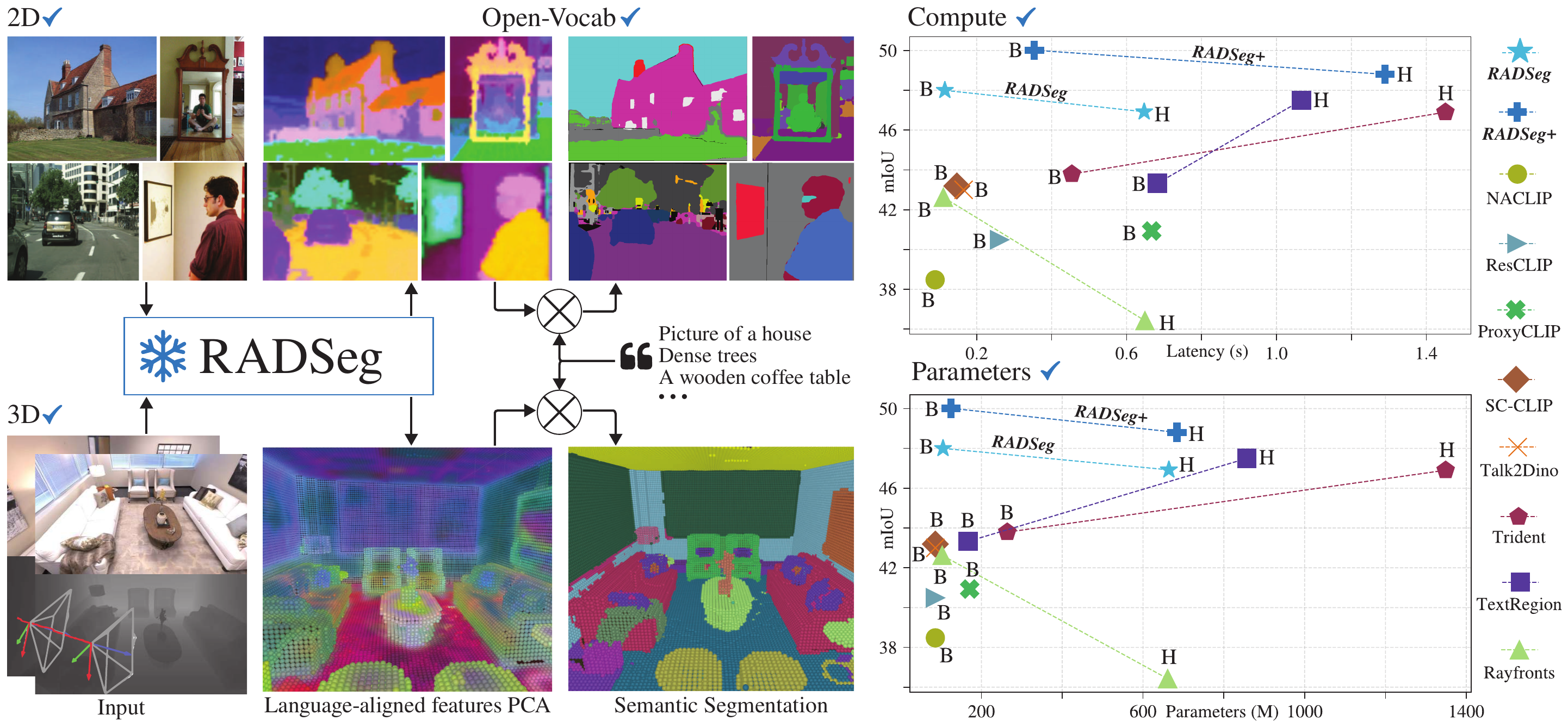}
\end{tabular}
\caption{\methodname{} is a dense, language-aligned feature encoder that enables low-parameter, low-latency open-vocabulary semantic segmentation in 2D and 3D. The efficiency plots report average latency, vision-backbone parameter count, and mIoU across five 2D datasets on a V100. By enhancing spatial locality of RADIO features, \textbf{\methodname{} outperforms previous state-of-the-art methods in accuracy while remaining highly efficient in terms of parameter counts and inference speed.}}
\label{fig:fig1}
\vspace{4mm}
\setcounter{figure}{1}
}
\makeatother
\else

\makeatletter
\let\@oldmaketitle\@maketitle
\renewcommand{\@maketitle}{\@oldmaketitle
\vspace{-7mm}

\centering
\captionsetup{type=figure, singlelinecheck=false}
\begin{tabular}{cccc}
\includegraphics[width=\textwidth]{figures/abstract_figure_compressed.pdf}
\end{tabular}
\vspace{-5mm}
\caption{\methodname{} is a dense, language-aligned feature encoder that enables low-parameter, low-latency open-vocabulary semantic segmentation in 2D and 3D. The efficiency plots report average latency, parameter count, and mIoU across five 2D datasets on a V100. By enhancing spatial locality of RADIO features, \textbf{\methodname{} outperforms previous state-of-the-art methods in accuracy while remaining highly efficient in terms of parameter counts and inference speed.}}
\label{fig:fig1}
\vspace{4mm}
\setcounter{figure}{1}
}
\makeatother

\fi

\maketitle
\begin{abstract}
Open-vocabulary semantic segmentation (OVSS) underpins many vision and robotics tasks that require generalizable semantic understanding. Existing approaches either rely on limited segmentation training data, which hinders generalization, or apply zero-shot heuristics to vision-language models (e.g CLIP), while the most competitive approaches combine multiple models to improve performance at the cost of high computational and memory demands. In this work, we leverage an overlooked agglomerative vision foundation model, RADIO, to improve zero-shot OVSS along three key axes simultaneously: mIoU, latency, and parameter efficiency. We present the first comprehensive study of RADIO for zero-shot OVSS and enhance its performance through self-correlating recursive attention, self-correlating global aggregation, and computationally efficient RADIO SAM mask refinement. Our approach, \methodname{}, achieves \textbf{6-30\% mIoU improvement} in the base ViT class while being \textbf{3.95x faster} and using \textbf{2.5x fewer parameters}. Surprisingly, \methodname{}-base \textbf{(106M)} outperforms previous combinations of huge vision models \textbf{(850-1350M)} in mIoU, achieving state-of-the-art accuracy with substantially lower computational and memory cost.
\end{abstract}

\vspace{-1em}
\section{Introduction}
\label{sec:intro}
Semantic segmentation models often need to be deployed in open-world environments and support an open-set of tasks. This has led to growing interest in open-vocabulary semantic segmentation (OVSS), which enables segmentation based on arbitrary natural langauge descriptions, with applications ranging from medical analysis \cite{luo2024brain}, robotic manipulation \cite{rashid2023language}, to autonomous exploration and navigation \cite{yokoyama2024vlfm, kim2025raven}. While vision language models (VLMs) such as CLIP \cite{radford2021clip} support open-vocabulary image classification, their reliance on global image-text encoding fundamentally limits their capacity for fine-grained, pixel-level localization required for OVSS. Various approaches have since attempted to adapt CLIP to the OVSS task. Some adopt a training-based approach, either incorporating CLIP into a new architecture \cite{li2022lseg, zhou2022maskclip+, xu2023san}, or fine-tuning CLIP specifically for OVSS \cite{rao2022denseclip}. Both approaches, however, suffer from limited generalization, as existing semantic segmentation datasets \cite{ade20k, caesar2018cocostuff, cityscapes, mottaghi2014context, voc} are orders of magnitude smaller than image-captioning datasets containing billions of image-text pairs \cite{schuhmann2022laion}. 

Training-free, zero-shot approaches have emerged as a promising alternative, alleviating the limitations of training-based methods. Dense-inference strategies feed multiple image crops into the global CLIP encoder to produce dense, language-aligned feature maps \cite{wysoczanska2024clipdiy, conceptfusion}, but incur substantial computational overhead. Alternative methods attempt to restore the spatial alignment of the CLIP's patch-wise tokens via lightweight modifications but yield limited accuracy \cite{hajimiri2025naclip, yang2025resclip, wang2024sclip}. More recent approaches \cite{shi2024trident, xiao2025textregion, lan2024proxyclip} improve segmentation by augmenting VLMs with combinations of other foundation models such as DINOv2 \cite{oquab2023dinov2} and SAM \cite{kirillov2023sam}, albeit at the cost of significantly increased model complexity, inference latency, and GPU memory overhead.

Given these limitations of existing OVSS methods, we investigate an underexplored agglomerative vision foundation model, RADIO \cite{ranzinger2024amradio}. By distilling knowledge from multiple foundation models into a single model, RADIO has shown strong performance across various vision benchmarks, in terms of generalization and efficiency. However, its potential for zero-shot OVSS remains largely unexplored. For instance, RayFronts \cite{alama2025rayfronts} demonstrated RADIO's zero-shot capability for 3D semantic mapping, but evaluations of its broader OVSS abilities remain limited.

In this work, we provide the first comprehensive study of RADIO for zero-shot OVSS. Our analysis includes compute and parameter efficiency for competing baselines, revealing sharp trade-offs between mIoU, inference latency, and GPU memory, illustrated in \cref{fig:fig1}. We also examine the ability of 2D OVSS methods to generate open-vocabulary semantic 3D maps, evaluating multi-view consistency.

Furthermore, we propose a novel and efficient pipeline, \methodname{}, based on RADIOv3. \methodname{} exploits an emergent property that enables alignment of RADIO's patch-wise features with language, enhances spatial locality by using Self-Correlating Recursive Attention (SCRA), and mitigates sliding-window artifacts through Self-Correlating Global Aggregation (SCGA).  Finally, we leverage RADIOv3's efficient access to SAM-huge features to further refine the obtained masks using only 0.8\% of SAM-huge parameters. \methodname{} achieves \textbf{6-30\% mIoU improvement} in the base ViT class over the next best baseline (Trident \cite{shi2024trident}) across datasets, while being \textbf{3.95x faster} and requiring \textbf{2.5x fewer parameters}. Remarkably, \methodname{}-base \textbf{(106M)} surpasses the mIoU of previous compositions of huge vision models like Trident \cite{shi2024trident} \textbf{(1350M)} and TextRegion \cite{xiao2025textregion} \textbf{(850M)}, \textbf{achieving state-of-the-art mIoU with low computational and GPU memory requirements}.

\noindent Our contributions can be summarized as:
\begin{itemize}
    \item We conduct the first thorough empirical study of the foundational model RADIO for zero-shot open-vocabulary semantic segmentation (ZSOVSS), demonstrating its superiority to other backbones.
    \item We introduce a novel pipeline using RADIOv3 that advances the state-of-the-art in ZSOVSS while maintaining low parameter counts and computational efficiency.
    \item We perform comprehensive evaluations on five 2D and three 3D datasets across various resolutions and model sizes, showing that \methodname{} consistently outperforms baselines at any resolution or model size budget.
\end{itemize}
\section{Related work}
\label{sec:related_work}

\begin{figure*}[htbp]
\centering
\includegraphics[width=0.85\textwidth]{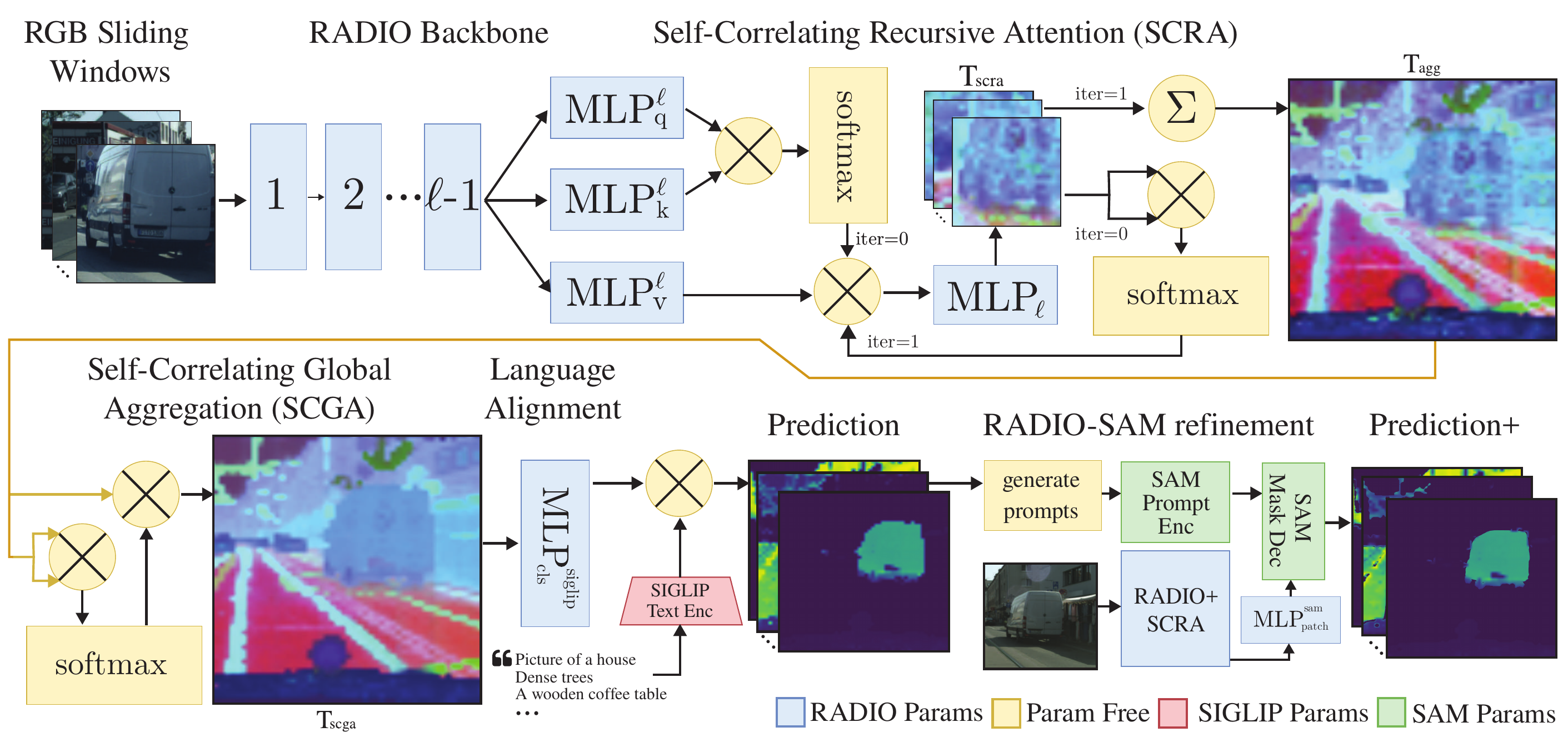}
\caption{Overview of the \methodname{} pipeline. RGB sliding windows are processed by the RADIO backbone. Self-Correlating Recursive Attention (SCRA) computes a similarity matrix from these outputs, which is recursively fed back into the last attention block of RADIO. Feature windows are aggregated into a feature map and refined through Self-Correlating Global Aggregation (SCGA) to reduce noise and windowing artifacts. Features are language-aligned with the SigLIP CLS token adaptor, and predictions are made by comparing them with text embeddings. Optionally, masks can be further refined using RADIO-SAM, requiring only \textbf{+13M} additional parameters.}
\label{fig:method}
\end{figure*}

\subsection{Vision and Language Foundation Models}
Vision-language foundation models (VLMs) trained on large amounts of image-text pairs have shown great generalization capabilties for various vision-language tasks \cite{radford2021clip,zhai2023siglip, tschannen2025siglip2}. VLMs align natural language and images into the same latent feature space, enabling zero-shot open-vocabulary classification by comparing text and image embeddings. However, such models are trained to align only a single CLS token that captures the global image context, hence termed global image-level VLMs. Consequently, the patch-wise/spatial tokens often exhibit poor spatial locality and language alignment. Recent research has sought to address these limitations to unlock open-vocabulary semantic segmentation (OVSS), which we discuss next.

\subsection{Open-Vocabulary Semantic Segmentation}
OVSS allows users to segment any image by simply providing natural language queries, removing the need for retraining when new categories are encountered in the wild. To achieve this, many existing methods fine-tune global image-level VLMs such as CLIP \cite{radford2021clip} on limited segmentation datasets at the expense of generalization ability 
\cite{li2022lseg, zhou2022maskclip+, xu2023san,rao2022denseclip}, i.e., training-based approaches. 

In contrast, training-free approaches have attracted particular interest, as they generate dense, class-agnostic, language-aligned feature maps directly from images and generalize to arbitrary datasets in a zero-shot manner. Some of these methods adopt a `dense-inference' strategy \cite{wysoczanska2024clipdiy, conceptfusion}, forwarding multiple crops or segments of an image through global VLMs and aggregating the results to obtain the map, yet incur substantial computational overhead. Others improve spatial locality and language alignment in global VLMs through lightweight attention modifications \cite{hajimiri2025naclip, wang2024sclip}, or residual connections \cite{yang2025resclip, bai2024scclip}. While these methods are computationally efficient, they yield lower segmentation accuracy. Another line of training-free methods augment VLMs with additional foundation models to produce dense language aligned features. ProxyClip \cite{lan2024proxyclip} leverages DINO's \cite{caron2021dino} strong spatial locality to refine CLIP features, Trident \cite{shi2024trident} extends it by incorporating SAM \cite{kirillov2023sam} for refinement; and TextRegion \cite{xiao2025textregion} integrates SAM2 \cite{ravi2024sam2} with CLIP. While these methods leveraging multiple foundation models achieve impressive segmentation accuracy, they incur significant memory and computational costs. 

Distinctively, our work incorporates RADIO \cite{ranzinger2024amradio,heinrich2025radio25}--an agglomerative model that distills knowledge from SAM, CLIP, SigLIP \cite{zhai2023siglip}, and DINOv2 \cite{oquab2023dinov2}--within a novel pipeline. By leveraging this unified foundation model, our approach outperforms the methods that explicitly combine multiple foundation models in segmentation accuracy, while maintaining the computational and memory efficiency of lightweight attention modification methods.

\subsection{3D Open Vocabulary Semantic Segmentation}
A growing body of work attempts to leverage 2D VLMs to achieve training-free/generalizable OVSS in 3D and create open-vocabulary 3D maps. These maps take various forms, ranging from semantic neural radiance fields (NeRFs) \cite{kim2024garfield, shafiullah2022clipfield}, gaussian splats \cite{jun2025drsplat, shi2024languagesplat}, scene graphs \cite{gu2024conceptgraphs, werby2024hovsg} to simple point clouds or voxels \cite{alama2025rayfronts, conceptfusion}. The most common strategy for obtaining dense, language-aligned features is to detect bounding boxes with GroundingDINO \cite{liu2023groundingdino}, use the detections to prompt SAM \cite{kirillov2023sam} for segmentation, then cropping each segment and forwarding it through CLIP \cite{radford2021clip}. However, this pipeline is computationally and memory intensive. Recently, RayFronts \cite{alama2025rayfronts} showed that using RADIOv2.5's SigLIP CLS-token adaptor to project RADIO's patch-wise tokens, combined with neighbor-aware attention \cite{hajimiri2025naclip}, can yield dense, language-aligned feature maps--achieving impressive results on 3D OVSS datasets. Building on this insight, we conduct the first comprehensive evaluation of RADIOv2.5 and RADIOv3 on both 2D and 3D ZSOVSS, and introduce novel spatial locality improvements to enable fast, memory efficient open-vocabulary 3D mapping with state-of-the-art segmentation performance.

\section{Method}
\label{sec:method}
This section overviews the \methodname{} pipeline (\cref{fig:method}), covering preliminaries on RADIO (Sec.~\ref{radio-prelim}), then \methodname{}'s main components--dense language alignment (Sec.~\ref{sec:dense-lang}), SCRA (Sec.~\ref{sec-scra}), SCGA (Sec.~\ref{sec-scga})--and the optional mask refinement process in \methodname+ (Sec.~\ref{sec-radio-sam}).

\subsection{RADIO Preliminary}

\label{radio-prelim}
RADIO \cite{ranzinger2024amradio, heinrich2025radio25} is an agglomerative vision foundation model that unifies knowledge distilled from CLIP \cite{radford2021clip}, SigLIP\cite{zhai2023siglip}, SAM \cite{kirillov2023sam}, and DINOv2 \cite{oquab2023dinov2} into a single model. RADIO employs a vision transformer (ViT) \cite{dosovitskiy2020vit}: an image $I \in \mathbb{R}^{H\times W \times 3}$ is divided into $N_{\text{patch}}=\frac{H\times W}{P_h\times P_w}$ non-overlapping patches $P \in \mathbb{R}^{N_{\text{patch}}\times P_h\times P_w\times 3}$, which are linearly projected into patch-wise tokens $T_{\text{patch}} \in \mathbb{R}^{N_{\text{patch}} \times D}$, where $D$ is the token embedding dimension. RADIO also prepends four separate CLS tokens, one for each teacher model (CLIP, SigLIP, SAM, DINOv2),  yielding $T_{\text{cls}} \in \mathbb{R}^{4 \times D}$. We denote the combined tokens as $T \in \mathbb{R}^{N \times D}$, where $N = N_{\text{patch}} + 4 + R$ ($R$ register tokens $T_{\text{reg}}$ are additionally introduced to mitigate high-norm non-local output tokens \cite{darcet2023vitreg}). Patch tokens $T_{\text{patch}}$ model spatially covariant/local features and CLS tokens $T_{\text{cls}}$ model global features summarizing the image. Unlike other agglomerative models that only align patch-wise output tokens to teachers (e.g. Theia \cite{shang2024theia}), RADIO aligns both patch-wise and CLS output tokens to their teacher counterparts through lightweight two-layer MLPs ($\text{MLP}_{\text{patch}}$ and $\text{MLP}_{\text{cls}}$), yielding adapted tokens $\tilde{T}_{\text{patch}}$ and $\tilde{T}_{\text{cls}}$. Formally,
\begin{equation}
    \tilde{T}_{\text{patch}}^{(m)} = \text{MLP}_{\text{patch}}^{(m)}(T_{l,\text{patch}}),
    \tilde{T}_{\text{cls}}^{(m)} = \text{MLP}_{\text{cls}}^{(m)}(T_{l,\text{cls}}^{(m)})
\end{equation}
where $m \in \{\text{clip}, \text{siglip}, \text{sam}, \text{dinov2}\}$\footnote{A notable exception is that CLIP patch-wise and CLS outputs have different embedding dimensions as the CLS output token goes through another projection to be aligned to language.}  and the subscript $l$ denote \textit{last} block for output tokens. Finally, RADIO demonstrates strong spatial locality in its patch-wise features, a property important for dense prediction tasks such as semantic segmentation. Since VLMs like CLIP and SigLIP only align the CLS tokens to language, RADIO learns a global image-language alignment.

\subsection{Getting Dense Language Alignment}
\label{sec:dense-lang}
RADIO has only been trained to align its SigLIP CLS output token $T^{\text{siglip}}_{l,\text{cls}}$ to the CLS token of the SigLIP teacher: 
\begin{equation}
\tilde{T}^{(\text{siglip})}_{\text{cls}} = \text{MLP}_{\text{cls}}^{(\text{siglip})}\big(T^{(\text{siglip})}_{l,\text{cls}}\big),
\min_{\theta} \mathcal{L}\big(\tilde{T}^{(\text{siglip})}_{\text{cls}}, \hat{T}^{(\text{siglip})}_{l,\text{cls}}\big)
\end{equation}
where $\tilde{T}$ are adapted tokens, $\hat{T}$ are original teacher tokens, and $\mathcal{L}(x,y)$ is a distance loss function that gets minimized.
This training only aligns the \textit{global summary} of an image with language, enabling \textbf{zero-shot open-vocabulary classification}. Similarly, RADIO aligns its patch output tokens $T_{l,\text{patch}}$ to SigLIP's output patch tokens $\hat{T}^{\text{siglip}}_{l,\text{patch}}$ through an MLP. However, SigLIP patch tokens exhibit poor spatial-language alignment as shown in \cref{tab:backbone_adaptor_ablations}. Consequently, RADIO trained a linear probe on small segmentation datasets, limiting generalization to a closed vocabulary. 

In contrast, we build on an observed emergent property of RADIO that enables dense, language-aligned feature extraction \cite{alama2025rayfronts}. By simply applying the SigLIP \textbf{CLS} token adaptor on RADIO's \textbf{patch} tokens, $$\text{MLP}_{\textbf{cls}}^{(\text{siglip})}(T_{l,\textbf{patch}}),$$ we obtain dense, language-aligned features suitable for OVSS. We empirically show that this emergent property holds for RADIOv2.5 and the newer RADIOv3, across model sizes and for both SigLIP and CLIP adaptation. However, directly using the resulting feature maps yields suboptimal segmentation performance, motivating further alignment refinements in the following sections.

\begin{figure}[t]
\centering
\includegraphics[width=0.9\linewidth]{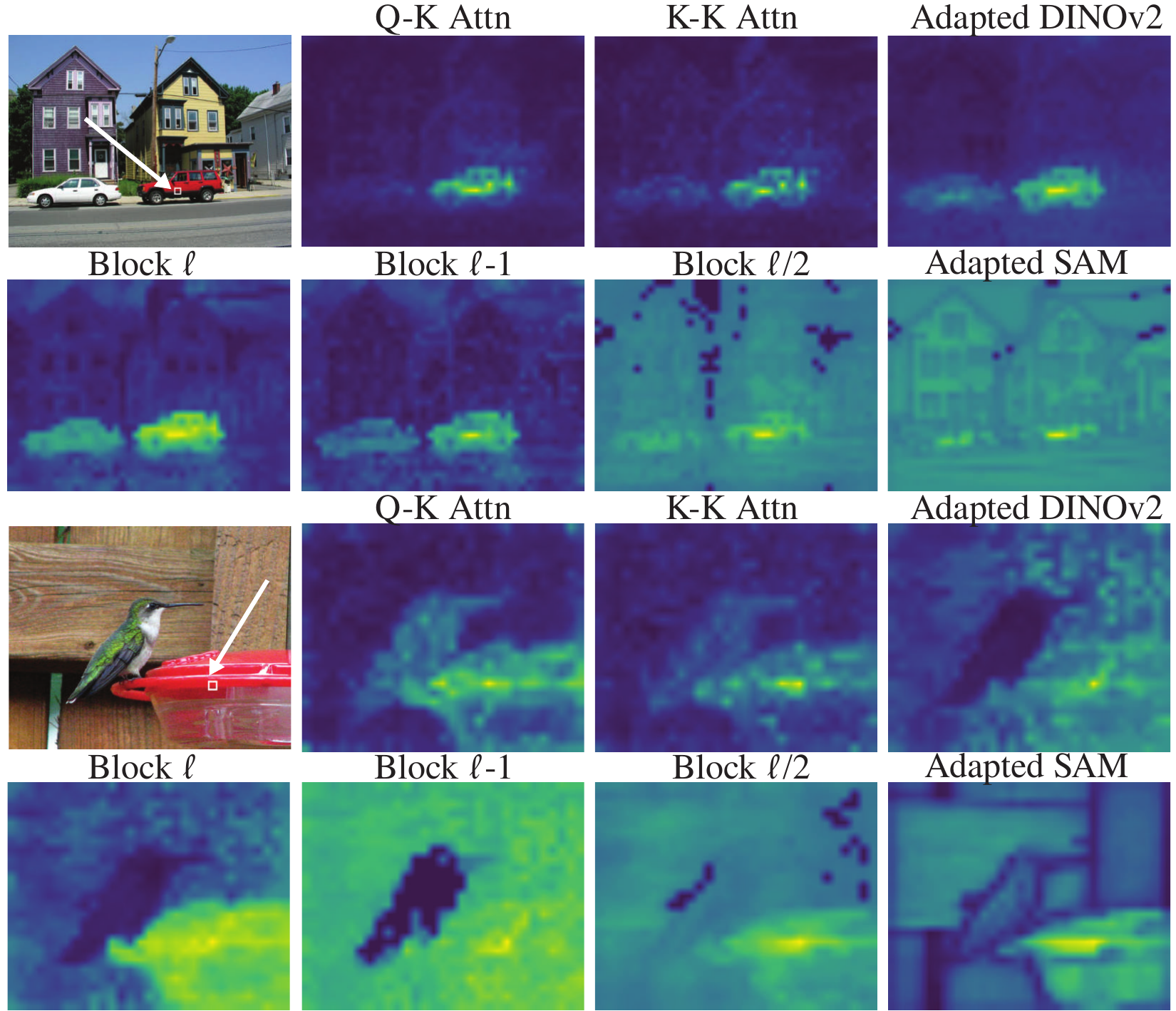}
\caption{Qualitative comparison of last block attention and patch-wise similarity at different parts of the RADIO framework. The output of the RADIO backbone (Block $l$) can consistently attend to semantically similar patches, motivating our SCRA approach.}
\label{fig:simattention}
\end{figure}

\subsection{Self-Correlating Recursive Attention}
\label{sec-scra}
A paradigm that has proven successful in previous training-free OVSS approaches \cite{wang2024sclip, hajimiri2025naclip, bai2024scclip, shi2024trident, yang2025resclip, lan2024proxyclip} is to exploit the last self-attention block, denoted as $\text{SA}_l$:
\begin{equation}
\text{SA}_l(T) = \text{Attn}(\text{MLP}^l_Q(T), \text{MLP}^l_K(T), \text{MLP}^l_V(T))
\end{equation}
, where $
    \text{Attn}(Q,K,V) = \text{softmax}\Big(\frac{QK^\top}{\sqrt{d_k}}\Big)V$. The last self-attention block in encoders like CLIP and SigLIP compromises spatial locality by focusing on CLS token. This can be mitigated by encouraging attention to semantically similar patches. Several methods have proposed to recover this spatial alignment either through light-weight attention modifications yielding limited accuracy, or by incorporating additional backbones to provide better attention, achieving increased accuracy at the cost of memory and latency.

In contrast to these extremes, our goal is to achieve high accuracy \textit{\textbf{without}} additional overheads. To this end, we examine what can be leveraged directly within  RADIO. As shown in \cref{fig:simattention}, we compare various attention and patch-similarity maps--computed without external backbones, including standard query-key and key-key attention, similarity maps from RADIO's intermediate blocks, and similarity maps derived from adapted DINOv2 and SAM patch tokens $\tilde{T}^{(\text{dinov2})}_{\text{patch}}$, $\tilde{T}^{(\text{sam})}_{\text{patch}}$. We find that RADIO's last block (Block $l$) offers the most stable spatial locality, providing a strong foundation for efficient zero-shot segmentation without external augmentations.

 Building on these findings, we propose a novel attention enhancement method, called \textbf{self-correlating recursive attention (SCRA)}. As shown in \cref{fig:method}, SCRA takes the output tokens $T_l\in\mathbb{R}^{N\times D}$ of the model, and computes a normalized correlation (cosine similarity) matrix 
$C = \check{T}_l \check{T}_l^{\top}$, where 
$\check{T}_l = T_l / \lVert T_l \rVert_2$. Cells with negative correlation between patches are set to $-\infty$, forcing patches to attend only to similar ones \cite{lan2024proxyclip}. More formally, 
\begin{align}
T_l &= \mathrm{MLP}_l\left(\mathrm{SA}\left(T_{l-1}\right)\right), \quad \check{T}_l = T_l / \lVert T_l \rVert_2 \\
C_{ij} &=
\begin{cases}
\check{T}_l^{(i)} \cdot \check{T}_l^{(j)}, 
& \text{if } \check{T}_l^{(i)} \cdot \check{T}_l^{(j)} \ge 0 \\
-\infty, & \mathrm{otherwise}
\end{cases} \label{eq:sim_mat}
\end{align}
A softmax is then applied across each column of a scaled $C$ to ensure that each patch's attention sums to one, keeping the resulting features within the convex hull of the originals. Finally, the last block's activations are recomputed using these self-correlation weights. SCRA computes its output $T_{\text{scra}}$ using RADIO's last block output $T_l$ as follows\footnotemark[1]\footnotetext[1]{Ignoring standard ViT layer norms, multiple attention heads, and residual connections for simplicity.}:
\begin{equation}
T_{\mathrm{scra}} = \mathrm{MLP}_l\left(\mathrm{softmax}(\tau_{\text{scra}} C) \cdot \mathrm{MLP}^l_V(T_{l-1})\right) \label{eq:scra}
\end{equation}
where $\tau_{\text{scra}} > 1$ is a temperature scaling factor that sharpens attention to only very similar patches. SCRA is able to improve the spatial-locality of RADIO features without introducing additional parameters and with minimal computational overhead of computing \cref{eq:sim_mat} and \cref{eq:scra}.

\subsection{Self-Correlating Global Aggregation}
\label{sec-scga}
When performing sliding window inference to produce the aggregated features map $T_{agg}$, we observe that the features describing the same entity yet in distinct windows can slightly differ, which introduces windowing artifacts \cite{shi2024trident} demonstrated in \cref{fig:method}. We propose a simple and effective \textbf{self-correlating global aggregation (SCGA)}. SCGA averages semantically similar features forcing cross-window consistency. We compute a self-correlation matrix $G = \check{T}_{\text{agg}} \check{T}_{\text{agg}}^{\top}$, where 
$\check{T}_{\text{agg}} = T_{\text{agg}} / \lVert T_{\text{agg}} \rVert_2$. Similar to SCRA, cells with negative correlation between patches are set to $-\infty$, a softmax is applied to each column of mean centered and scaled $G$. The correlation matrix $G$ is used as attention to aggregate the tokens:
\begin{align}
G_{ij} &=
\begin{cases}
\check{T}_{\text{agg}}^{(i)} \cdot \check{T}_{\text{agg}}^{(j)}-\mu_G, 
& \text{if } \check{T}_{\text{agg}}^{(i)} \cdot \check{T}_{\text{agg}}^{(j)} \ge 0 \\
-\infty, & \mathrm{otherwise}
\end{cases} \\
T_{\mathrm{\text{scga}}} &=\mathrm{softmax}(\tau_{\text{scga}} G) \cdot T_{\text{agg}} 
\end{align}
Qualitatively, \cref{fig:method} shows how  SCGA suppresses the noise and windowing artifacts, while maintaining the sharpness of the feature map, all without additional backbones.

Together, dense language alignment, SCRA, and SCGA constitute the core components of \methodname{}. In the next subsection, we introduce an optional mask-refinement stage that extends the framework to \methodname\textbf{+}.

\begin{figure*}[!htbp]
\centering
\includegraphics[width=0.95\textwidth]{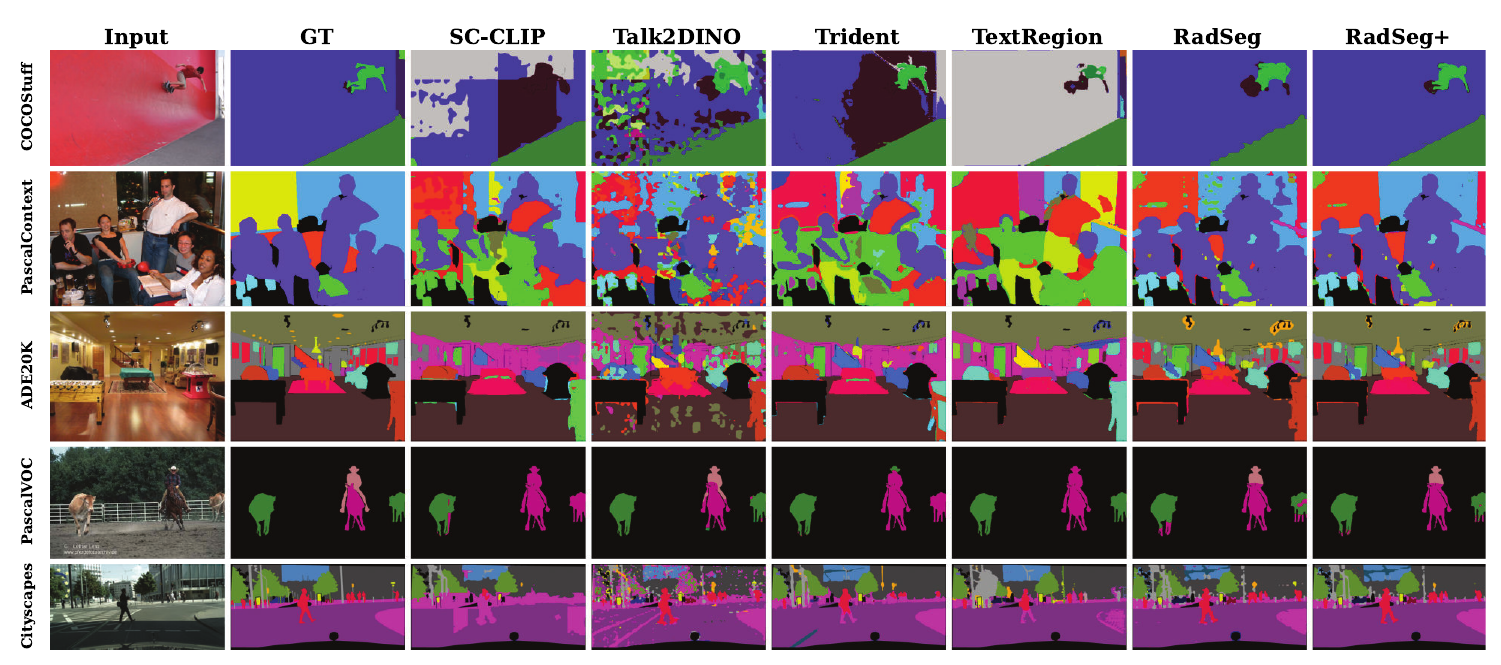}
\caption{\textbf{Qualitative 2D Open-Vocabulary Semantic Segmentation Results.} For each of the five benchmark datasets, we show a representative example and compare \methodname{} and \methodname{}\textbf{+} with competitive baselines (SC-CLIP, Talk2DINO, Trident, and TextRegion). Both \methodname{} and \methodname{}\textbf{+} produce noticeably clearer and more accurate segmentation maps across all cases.}
\label{fig:2d_qualitative_results}
\end{figure*}

\subsection{RADIO-SAM Refinement}

\label{sec-radio-sam}
Most OVSS methods rely on ViT backbones, producing coarse feature maps with 14-16x downsampling. Even with overlapping windows, this resolution gap remains. Prior work closes this gap either with RGB-based refiners such as PAMR \cite{araslanov2020pamr}, which are computationally heavy, or with SAM-based refinement \cite{shi2024trident, lin2025samrefiner}, which improves quality but requires an additional backbone. 

Within the RADIO framework, however, we can \textbf{leverage SAM-huge at only a +13M increase in parameters}. RADIO readily provides an adaptor that maps its output patch tokens $T_{l,\text{patch}}$ into the SAM-huge token space $
\tilde{T}_{\text{patch}}^{(\text{sam})}
    = \mathrm{MLP}_{\text{patch}}^{(\text{sam})}\!\left(T_{l,\text{patch}}\right).
$
As a result, we only need the decoder, neck, and prompt encoders of SAM-huge, which constitute \textbf{0.8\% of the total SAM-huge parameters}.

Following SAM-based refinement strategies \cite{shi2024trident, xiao2025textregion}, we generate point, box, and mask prompts from the coarse predictions of $T_{\text{scga}}$. Rather than simply reusing $T_{\text{scga}}$ features, which we find degrades accuracy, we re-encode the full image with RADIO+SCRA (no sliding windows) to feed into SAM. As shown in \cref{fig:method},  RADIO-SAM refinement sharpens boundaries and yields better final masks.

\section{Experiments}
\label{sec:experiments}

\subsection{2D Zero-Shot Open Vocabulary Segmentation}
\label{sec:experiments_2d}

\begin{table*}[!t]
\centering
\caption{2D zero-shot open-vocabulary semantic segmentation mIoU results across resolution limits and model sizes. Ranking shown as \colorbox{green!25}{\textbf{first}}, \colorbox{yellow!30}{second}, and \colorbox{orange!30}{third}. One ranking for base and huge is reported to highlight that \methodname{}-base is also superior to huge baselines. \underline{Underlined} cells show results obtained at lower resolution. Numbers at specific resolutions are included in the supplementary.}
\resizebox{\linewidth}{!}{
\setlength{\tabcolsep}{3.5pt}
\begin{tabular}{l|cccccc|cccccc|cccccc}
\hline
\multirow{2}{*}{\textbf{Methods}} &
\multicolumn{6}{c|}{\textbf{$\leq$ Low Resolution}} &
\multicolumn{6}{c|}{\textbf{$\leq$ Mid Resolution}} &
\multicolumn{6}{c}{\textbf{$\leq$ High Resolution}} \\
& CTX & VOC & Stuff & ADE & City & Avg
& CTX & VOC & Stuff & ADE & City & Avg
& CTX & VOC & Stuff & ADE & City & Avg \\
\hline
\multicolumn{19}{c}{\textbf{Base Models}} \\
\hline
NACLIP \cite{hajimiri2025naclip}     & 35.17 & 79.71 & 23.30 & 17.42 & 35.49 & 38.22 & \underline{35.17} & 80.29 & 23.64 & 17.78 & 35.98 & 38.57 & \underline{35.17} & \underline{80.29} & \underline{23.64} & 17.80 & 36.74 & 38.73 \\
ResCLIP \cite{yang2025resclip}    & 36.80 & 82.32 & 24.70 & 18.03 & 35.85 & 39.54 & \underline{36.80} & 85.89 & 25.07 & 18.55 & 36.19 & 40.50 & \underline{36.80} & \underline{85.89} & \underline{25.07} & 18.57 & 36.88 & 40.64 \\
RayFronts \cite{alama2025rayfronts}  & 36.81 & 72.59 & 25.34 & 23.38 & 36.29 & 38.88 & 38.07 & 80.12 & 27.39 & 24.63 & 40.47 & 42.14 & \underline{38.07} & \underline{80.12} & \underline{27.39} & \underline{24.63} & \underline{40.47} & 42.14 \\
ProxyCLIP \cite{lan2024proxyclip}   & 38.74 & 78.18 & 26.11 & 19.71 & 39.69 & 40.49 & \underline{38.74} & 80.31 & 26.41 & \underline{19.71} & 40.39 & 41.11 & \underline{38.74} & \underline{80.31} & \underline{26.41} & \underline{19.71} & \underline{40.39} & 41.11 \\
SC-CLIP \cite{bai2024scclip}   & 40.12 & 84.29 & 26.62 & 20.06 & 41.02 & 42.42 & \underline{40.12} & 87.67 & 27.25 & 20.68 & \underline{41.02} & 43.35 & \underline{40.12} & \underline{87.67} & \underline{27.25} & \underline{20.68} & 41.24 & 43.39 \\
Talk2Dino \cite{barsellotti2024talk2dino} & 39.43 & 85.68 & 27.43 & 20.42 & 38.05 & 42.20 & 40.31 & \underline{85.68} & 27.89 & 21.67 & 39.47 & 43.00 & \underline{40.31} & \underline{85.68} & \underline{27.89} & 21.70 & 41.15 & 43.35 \\
Trident  \cite{shi2024trident}   & 41.62 & 83.74 & 28.22 & 21.17 & 41.86 & 43.32 & 42.16 & 84.50 & 28.24 & 21.98 & 42.08 & 43.79 & \underline{42.16} & \underline{84.50} & \underline{28.24} & \underline{21.98} & 42.19 & 43.81 \\
TextRegion \cite{xiao2025textregion} & 41.53 & 83.83 & 27.39 & 21.21 & 40.49 & 43.17 & 42.29 & 84.21 & \underline{27.39} & 21.74 & 41.26 & 43.38 & 42.88 & \underline{84.21} & 28.62 & 22.55 & 42.15 & 44.08 \\
\methodname{} & \cellcolor{orange!30} \textbf{44.49} & \textbf{87.24} & \textbf{29.79} & \cellcolor{orange!30} \textbf{27.16} & \textbf{42.04} & \textbf{46.14} & \cellcolor{orange!30} \textbf{45.64} & \textbf{89.28} & \cellcolor{yellow!30} \textbf{30.76} & \cellcolor{orange!30} \textbf{28.96} & \textbf{45.35} & \cellcolor{orange!30} \textbf{48.00} & \textbf{\underline{45.64}} & \textbf{\underline{89.28}} & \textbf{\underline{30.76}} & \cellcolor{orange!30} \textbf{\underline{28.96}} & \cellcolor{orange!30} \textbf{48.79} & \cellcolor{orange!30} \textbf{48.69} \\
\methodname{}\textbf{+} & \cellcolor{green!30}{\textbf{48.24}} & {\textbf{88.46}} & \cellcolor{green!30}{\textbf{31.69}} & \cellcolor{green!30}{\textbf{29.63}} & \cellcolor{orange!30}{\textbf{45.58}} & \cellcolor{green!30}{\textbf{48.84}} & \cellcolor{green!30}{\textbf{48.48}} & \cellcolor{yellow!30}{\textbf{90.14}} & \cellcolor{green!30}{\textbf{32.48}} & \cellcolor{green!30}{\textbf{30.83}} & \cellcolor{green!30}{\textbf{48.10}} & \cellcolor{green!30}{\textbf{50.01}} & \cellcolor{green!30}{\textbf{\underline{48.48}}} & \cellcolor{yellow!30}{\textbf{90.39}} & \cellcolor{green!30}{\textbf{\underline{32.48}}} & \cellcolor{green!30}{\textbf{30.86}} & \cellcolor{green!30}{\textbf{50.96}} & \cellcolor{green!25}{\textbf{51.17}} \\

\hline
\multicolumn{19}{c}{\textbf{Huge Models}} \\
\hline
RayFronts \cite{alama2025rayfronts}  & 31.67 & 72.59 & 23.31 & 20.46 & 28.98 & 35.40 & 32.29 & 73.36 & 23.44 & 21.19 & 31.84 & 36.42 & \underline{32.29} & \underline{73.36} & \underline{23.44} & \underline{21.19} & 34.27 & 36.91 \\
ProxyCLIP \cite{lan2024proxyclip}  & 39.16 & 78.02 & 26.19 & 23.90 & 43.64 & 42.18 & \underline{39.16} & 83.03 & 27.76 & 24.05 & 43.92 & 43.58 & \underline{39.16} & \underline{83.03} & \underline{27.76} & \underline{24.05} & \underline{43.92} & 43.58 \\
Trident  \cite{shi2024trident}   & 43.16 & 87.97 & 28.55 & 25.64 & \cellcolor{yellow!30} 46.87 & 46.44 & 44.32 & 88.67 & \underline{28.55} & 26.70 & \cellcolor{orange!30} \underline{46.87} & 47.02 & \underline{44.32} & \underline{88.67} & \underline{28.55} & 27.02 & 47.34 & 47.18 \\
TextRegion \cite{xiao2025textregion} & 44.04 & \cellcolor{orange!30} 89.53 & \cellcolor{yellow!30} 30.19 & 24.53 & \cellcolor{green!25} 47.35 & \cellcolor{yellow!30} 47.13 & 44.76 & \underline{89.53} & \underline{30.19} & 26.42 & \cellcolor{yellow!30} \underline{47.35} & 47.65 & \cellcolor{orange!30} 46.13 & \underline{89.53} & \cellcolor{yellow!30} 31.22 & 27.30 & \underline{47.35} & 48.31 \\
\methodname{} & \textbf{42.27} & \cellcolor{yellow!30} \textbf{89.58} & \textbf{28.30} & \textbf{25.96} & \textbf{38.85} & \textbf{44.99} & \textbf{44.80} & \cellcolor{orange!30} \textbf{89.74} & \textbf{28.93} & \textbf{28.21} & \textbf{42.93} & \textbf{46.92} & \textbf{45.01} & \cellcolor{orange!30} \textbf{\underline{89.74}} & \textbf{29.46} & \textbf{\underline{28.21}} & \textbf{47.75} & \textbf{48.03} \\
\methodname{}\textbf{+} & \cellcolor{yellow!30}{\textbf{45.75}} & \cellcolor{green!30}{\textbf{90.39}} & \cellcolor{orange!30}{\textbf{30.17}} & \cellcolor{yellow!30}{\textbf{27.86}} & {\textbf{42.49}} & \cellcolor{yellow!30}{\textbf{47.73}} & \cellcolor{yellow!30}{\textbf{47.76}} & \cellcolor{green!30}{\textbf{90.44}} & \cellcolor{orange!30}{\textbf{30.50}} & \cellcolor{yellow!30}{\textbf{30.09}} & {\textbf{46.48}} & \cellcolor{yellow!30}{\textbf{49.38}} & \cellcolor{yellow!30}{\textbf{47.98}} & \cellcolor{green!30}{\textbf{\underline{90.44}}} & \cellcolor{orange!30}{\textbf{30.79}} & \cellcolor{yellow!30}{\textbf{\underline{30.09}}} & \cellcolor{yellow!30}{\textbf{50.58}} & \cellcolor{yellow!25}{\textbf{50.48}} \\
\hline
\end{tabular}
}
\label{tab:ovss_2d_lte}
\end{table*}

\textbf{Benchmark Settings.} Consistent with previous methods \cite{shi2024trident, xiao2025textregion, hajimiri2025naclip, barsellotti2024talk2dino}, we evaluate our method across five widely used 2D semantic segmentation benchmarks: PASCAL VOC \cite{voc}, PASCAL Context  \cite{mottaghi2014context}, COCO-Stuff \cite{caesar2018cocostuff}, Cityscapes \cite{cityscapes}, and ADE20K \cite{ade20k}. In prior work, different datasets and baselines were often evaluated at different resolutions, making it diffcult to disentangle performance gains from higher resolutions versus improvements due to the method itself. Thus, for a comprehensive evaluation, we evaluate three representative resolution standards set by previous methods: low (336-560) \cite{hajimiri2025naclip, wang2024sclip, yang2025resclip}, mid (336-688) \cite{shi2024trident}, and high (672-1344) \cite{xiao2025textregion}, with details per dataset in the supplementary. We do not allow any method, to access a higher input resolution than its setting, yet, it can operate on lower resolutions. We do not employ heavy post-processing (e.g PAMR \cite{araslanov2020pamr}) for any of the methods.

We follow standard OVSS protocols, each class is embedded in prompt templates, encoded with the text encoder, and used to obtain segmentation masks through cosine similarity with the feature maps. We ignore the ambiguous `background' label. We report mean Intersection over Union (mIoU) on the validation split of each dataset. 

We report our two methods, \methodname{} and \methodname{}\textbf{+}, without and with RADIO-SAM refinement. We evaluate both the base and huge variants of RADIOv3, using SigLIP2 as the feature adaptor with $\tau_{\text{scra}}=10, \tau_{\text{scga}}=10$.

\textbf{Baselines}. We compare our approach against a diverse set of SOTA training-free OVSS methods. This includes CLIP-based adaptations such as NACLIP \cite{hajimiri2025naclip}, ResCLIP \cite{yang2025resclip}, and SC-CLIP \cite{bai2024scclip}, which refine attention blocks within the CLIP architecture; hybrid multi-model fusion methods such as ProxyCLIP \cite{lan2024proxyclip}, Talk2DINO \cite{barsellotti2024talk2dino}, Trident \cite{shi2024trident}, and TextRegion \cite{xiao2025textregion}, which leverage multiple vision foundation models including CLIP, DINO, and SAM; and the only RADIO-based method Rayfronts \cite{alama2025rayfronts}.

\textbf{Quantitative and Qualitative Results.}
\cref{tab:ovss_2d_lte} shows how well methods perform within a resolution limit across datasets. \methodname{} establishes a clear performance margin over prior zero-shot OVSS methods. In the \textit{base} setting, \methodname{} achieves the highest average mIoU at all resolution budgets, outperforming the closest baselines by \textbf{+3--5\% mIoU}. \methodname{}\textbf{+} strengthens this lead further, improving average mIoU by up to \textbf{+7\% mIoU} over the next best baseline. Remarkably, our base model  already surpasses several huge-model baselines, underscoring the effectiveness of RADIO and our pipeline.

We show qualitative comparisons of \methodname{} and \methodname{}\textbf{+} against baselines, using base models, in \cref{fig:2d_qualitative_results}. \methodname{} demonstrates superior semantic consistency and more precise segmentation boundaries, especially evident in challenging categories like ``person" and in complex scenes in Pascal Context and Cityscapes.

\subsection{3D Zero-Shot Open Vocabulary Segmentation}
\label{sec:experiments_3d}

\begin{figure}[tbp]
\centering
\includegraphics[width=0.97\linewidth]{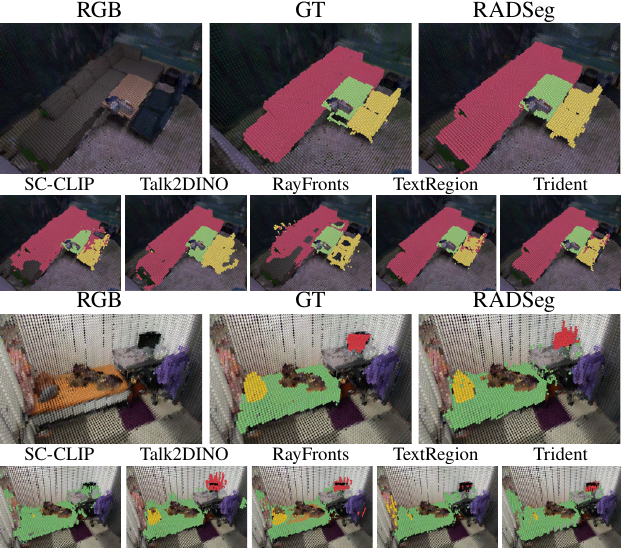}
\caption{\textbf{Qualitative 3D Open-Vocabulary Semantic Segmentation Results.} We show two scenes: one from Replica (``chair", ``table", ``couch" classes), and one from ScanNet++ (``bed", ``pillow", ``monitor" classes). Segmented voxels are overlaid on the RGB for visualization. Across all 3D baselines, \methodname{} provides more accurate segmentations with far fewer outlier voxels.}
\label{fig:3d_qualitative_results}
\vspace{-2mm}
\end{figure}

\begin{table*}[!t]
\centering
\caption{3D zero-shot open-vocabulary semantic segmentation mIoU results. Ranking shown as \colorbox{green!25}{\textbf{first}}, \colorbox{yellow!30}{second}, and \colorbox{orange!30}{third}. \methodname{} consistently shows superior performance across datasets emphasizing its multi-view consistency.}
\label{tab:semseg3d}
\resizebox{0.9\textwidth}{!}{
\begin{tabular}{lcccccccccccc}
\toprule
& \multicolumn{6}{c}{\textbf{Feature Space 3D Aggregation}} & \multicolumn{6}{c}{\textbf{Probability Space 3D Aggregation}} \\ 
\cmidrule(lr{0.75em}){2-7} \cmidrule(lr{0.75em}){8-13}

& \multicolumn{2}{c}{\textbf{Replica}~\cite{straub2019replica}} & \multicolumn{2}{c}{\textbf{ScanNet}~\cite{dai2017scannet}} & 
\multicolumn{2}{c}{\textbf{ScanNet++}~\cite{yeshwanth2023scannet++}} &
\multicolumn{2}{c}{\textbf{Replica}~\cite{straub2019replica}} & \multicolumn{2}{c}{\textbf{ScanNet}~\cite{dai2017scannet}} &
\multicolumn{2}{c}{\textbf{ScanNet++}~\cite{yeshwanth2023scannet++}} \\ 
\cmidrule(lr{0.75em}){2-3} \cmidrule(lr{0.75em}){4-5} 
\cmidrule(lr{0.75em}){6-7} \cmidrule(lr{0.75em}){8-9}
\cmidrule(lr{0.75em}){10-11} \cmidrule(lr{0.75em}){12-13}

{\textbf{Methods}} &
mIoU & f-mIoU & 
mIoU & f-mIoU & 
mIoU & f-mIoU & 
mIoU & f-mIoU & 
mIoU & f-mIoU & 
mIoU & f-mIoU \\ 
\midrule

SC-CLIP \cite{bai2024scclip} &
21.33 & 43.57 & 28.04 & 40.57 & 
22.03 & 42.83 &
21.98 & 44.31 & 28.59 & 40.89 
& 23.90 & 43.49 \\

Talk2Dino \cite{barsellotti2024talk2dino} & 
23.12 & 35.07 & 28.58 & 32.88 &
\cellcolor{orange!30}{28.91} & 39.88 &
23.94 & 36.86 & 29.13 & 34.47 &
29.54 & 41.37 \\

Trident \cite{shi2024trident} & 
23.62 & 47.18 & \cellcolor{orange!30}{32.23} & \cellcolor{orange!30}{43.12} &
26.44 & \cellcolor{orange!30}{49.68} &
25.44 & 47.48 & 32.78 & 43.67 & 
27.83 & 50.54 \\

Trident (w/refine) & 
-- & -- & -- & -- & -- & -- &
27.38 & 49.81 & 32.63 & 44.17 &
29.89 & 53.75 \\

TextRegion \cite{xiao2025textregion} & 
\cellcolor{orange!30}{27.15} & \cellcolor{orange!30}{50.50} & \cellcolor{yellow!30}{34.42} & \cellcolor{yellow!30}{44.47} &
27.64 & \cellcolor{yellow!30}{54.52} &
28.35 & 50.90 & \cellcolor{orange!30}{35.83} & \cellcolor{orange!30}{46.09} &
28.14 & \cellcolor{orange!30}{54.81} \\

RayFronts \cite{alama2025rayfronts} & 
\cellcolor{yellow!30}{28.39} & \cellcolor{yellow!30}{52.05} & 31.02 & 42.14 &
\cellcolor{yellow!30}{30.72} & 47.79 &
\cellcolor{orange!30}{31.28} & \cellcolor{orange!30}{53.28} & 31.05 & 42.30 &
\cellcolor{orange!30}{31.61} & 49.51 \\

\methodname{} & 
\cellcolor{green!25}{\textbf{32.29}} & \cellcolor{green!25}{\textbf{58.96}} &\cellcolor{green!25}{\textbf{36.29}} & \cellcolor{green!25}{\textbf{45.64}} &
\cellcolor{green!25}{\textbf{36.22}} & \cellcolor{green!25}{\textbf{55.73}} &
\cellcolor{yellow!30}{\textbf{32.68}} & \cellcolor{yellow!30}{\textbf{57.02}} &\cellcolor{yellow!30}{\textbf{36.41}} & \cellcolor{yellow!30}{\textbf{46.92}} &
\cellcolor{yellow!30}{\textbf{36.81}} & \cellcolor{yellow!30}{\textbf{57.14}} \\

\methodname{}\textbf{+} & 
-- & -- & -- & -- & -- & -- &
\cellcolor{green!25}{\textbf{33.05}} & \cellcolor{green!25}{\textbf{59.21}} &\cellcolor{green!25}{\textbf{38.58}} & \cellcolor{green!25}{\textbf{48.49}} &
\cellcolor{green!25}{\textbf{39.93}} & \cellcolor{green!25}{\textbf{59.25}} \\

\bottomrule
\end{tabular}}
\end{table*}

While 2D segmentation evaluates per-frame semantic understanding, it fails to capture multi-view semantic consistency, essential for embodied perception. Thus, we extend our evaluation to 3D, examining how different encoders generalize within a 3D reconstruction pipeline. We focus on evaluating encoders rather than 3D scene representations. Thus, we adopt a simple setup across methods, in which 2D outputs are unprojected onto point clouds and aggregated within voxels, averaging coincident point features.

\textbf{Benchmark Settings.} 
 We test two 2D-to-3D lifting strategies: projecting the features, or projecting the segmentation probabilities. Probability projection prevents feature collisions and supports post-segmentation refinement, yet forces closed-set outputs. Feature projections retain open-vocabulary flexibility at the cost of possible aliasing.

We evaluate on Replica \cite{straub2019replica} and ScanNet \cite{dai2017scannet} following prior work \cite{alama2025rayfronts}, and additionally on ScanNet++ \cite{yeshwanth2023scannet++}. For a fair comparison, we extend the top-performing 2D baselines Trident, SC-CLIP, TextRegion, Talk2DINO, and RayFronts, to 3D using the same projection and fusion procedure. Since 3D dense feature maps require much more GPU memory, we only evaluate base models for all baselines and our method.

\textbf{Quantitative and Qualitative Results.} \cref{tab:semseg3d} shows that \methodname{} establishes a clear performance margin over the baselines. Under feature-space aggregation, our approach achieves the highest mIoU on all the datasets, improving over the strongest baselines by \textbf{+3--5\% mIoU} on average. When evaluated in probability space, which allows for the evaluation of \methodname{}\textbf{+}, it again ranks first or second in every setting: without refinement, it secures the \textbf{second-highest} metrics across datasets, while with refinement it achieves SOTA results in terms of mIoU/f-mIoU with \textbf{33.05\%/59.21\%} on Replica, \textbf{38.58\%/48.49\%} on ScanNet, and \textbf{33.27\%/48.35\%} on ScanNet++. Notably, while RayFronts struggles in 2D, its use of RADIO allows it to be competitive in 3D, showing the importance of a complementary 3D evaluation. Overall, \methodname{} consistently outperforms CLIP-based, DINO-based, and multi-model baselines showing its strong multi-view consistency.

\cref{fig:3d_qualitative_results} presents qualitative comparisons between \methodname{} and all the baselines in our evaluation. Across a diverse set of queries, \methodname{} consistently yields cleaner and more semantically coherent predictions, producing significantly fewer visual outliers than competing methods. 

\subsection{Ablation Studies}
All ablations are conducted on 2D datasets using mid resolution.
\label{sec:experiments_ablation}

\begin{table}[!htbp]
\centering
\caption{Ablation of different RADIO language adapters across model sizes. Avg mIoU reported across 2D datasets. R=RADIO.}
\centering
\resizebox{0.725\linewidth}{!}{
\begin{tabular}{l|ccc}
\hline
Methods & Base & Large & Huge \\
\hline
Rv2.5-SigLIP$_{\text{cls}}$    & 39.96 & \textbf{38.55} & 36.97 \\
Rv2.5-SigLIP$_{\text{patch}}$     & 0.32 & 0.31 & 0.36 \\
Rv2.5-CLIP$_{\text{cls}}$    & 39.43 & 37.79 & 34.85 \\
Rv2.5-CLIP$_{\text{patch}}$    & NA & NA & NA \\
Rv3-SigLIP2$_{\text{cls}}$   & \textbf{44.62} & 0.86 & \textbf{41.24} \\
Rv3-SigLIP2$_{\text{patch}}$   & 0.73 & 0.76 & 0.25 \\
Rv3-CLIP$_{\text{cls}}$   & 42.32 & 0.75 & 36.79 \\
Rv3-CLIP$_{\text{patch}}$   & NA & NA & NA \\
\hline
\end{tabular}}
\label{tab:backbone_adaptor_ablations}
\end{table}

\begin{table}[!htbp]
\centering
\caption{\textbf{All proposed components yield significant improvments.} mIoU and latency on a V100 across 2D datasets is shown.}
\resizebox{\columnwidth}{!}{%
\begin{tabular}{l|cccccc|c}
\hline
\multirow{2}{*}{Methods} & \multicolumn{6}{c|}{mIoU (\%)} & \multirow{2}{*}{Lat (s)} \\

 & CTX & VOC & Stuff & ADE & City & Avg & \\
\hline
Base & 40.4 & 88.07 & 27.41 &  27.3 & 39.94  & 44.62 & 0.107 \\
+ SCRA & 43.07 & 88.03 & 29.87 & 27.69 & 44.5 &  46.63 & 0.111 \\
+ SCGA & 45.64 & 89.28 & 30.76 & 28.96 & 45.35 & 48.00 &  0.115 \\
+ R-SAM & \textbf{48.48} & \textbf{90.14} & \textbf{32.48} & \textbf{30.83} & \textbf{48.10} & \textbf{50.01} & 0.354 \\
\hline
\end{tabular}%
}
\label{tab:ablation_hier}
\end{table}

\newcommand{\gd}[1]{{\scriptsize\textcolor{green!60!black}{(#1)}}}
\newcommand{\bd}[1]{{\scriptsize\textcolor{orange!80!black}{(#1)}}}
\newcommand{\mioud}{\phantom{{\scriptsize(+0.0)}}}
\newcommand{\paramd}{\phantom{{\scriptsize(-66)}}}
\newcommand{\latd}{\phantom{{\scriptsize(-0.56)}}}
\begin{table}[htb]
\caption{\textbf{RADIO can unlock efficiency and mIoU gains for different baselines.} Average mIoU across 2D datasets, number of parameters for base models and latency on a V100. }
\centering
\resizebox{0.925\linewidth}{!}{
\begin{tabular}{l|lll}
\hline
Methods & mIoU (\%) & Params (M) & Lat (s) \\
\hline
NACLIP \cite{hajimiri2025naclip}        & 38.48\mioud & 86\paramd  & 0.089\latd \\
NARADIO                                & 44.54\gd{+6.1} & 106\bd{+20} & 0.107\bd{+0.02} \\
\specialrule{0.3pt}{2pt}{2pt}
ProxyCLIP \cite{lan2024proxyclip}      & 40.87\mioud & 171\paramd & 0.667\latd \\
ProxyRADIO-D                           & 45.44\gd{+4.6} & 106\gd{-65} & 0.124\gd{-0.54} \\
ProxyRADIO-S                           & 46.05\gd{+5.2} & 106\gd{-65} & 0.124\gd{-0.54} \\
\specialrule{0.3pt}{2pt}{2pt}
ResCLIP \cite{yang2025resclip}         & 40.33\mioud & 86\paramd  & 0.261\latd \\
ResRADIO                                & 45.18\gd{+4.9} & 106\bd{+20} & 0.123\gd{-0.14} \\
\specialrule{0.3pt}{2pt}{2pt}
\methodname{}                           & \textbf{48.00}\mioud & 106\paramd & 0.115\latd \\
\bottomrule
\end{tabular}
}
\label{tab:attention-ablation}
\end{table}

\textbf{Exploring RADIO's language alignment.} \cref{tab:backbone_adaptor_ablations} summarizes the performance of different RADIO language adapters. Following RADIO's training regimen of projecting RADIO patch features to SigLIP/CLIP patch features shows no language alignment with a dimensionality mismatch in case of CLIP, while using CLS adaptors on patch features yields high OVSS mIoU (34-44\%) across datasets, highlighting RADIO's emergent dense language alignment. RADIOv3 with SigLIP2 CLS shows the highest alignment which we adopt in our method. Notably, RADIO's language alignment excels with base model sizes, making it all the more suitable for efficient zero-shot OVSS.

\textbf{Effectiveness of \methodname{} components}.
\cref{tab:ablation_hier} summarizes 2D OVSS performance across datasets for evaluating our attention and refinement components. Notably, SCRA and SCGA provide consistent improvements with an average +2\%, and +1.4\% respectively, at negligible computational overhead. RADIO-SAM refinement improves by another +2\% yet slows down the pipeline  by 3$\times$. Regardless, \methodname{}+ (which includes RADIO-SAM refinement) remains significantly faster than competing methods.

\textbf{Can RADIO improve baseline approaches?} To demonstrate RADIO's broader applicability in OVSS beyond \methodname{}, we adapt some of the previous approaches that can readily benefit from the backbone. Specifically, we use RADIOv3-b with adapted SigLIP2 features and incorporate neighbor aware \cite{hajimiri2025naclip}, residual \cite{yang2025resclip}, and proxy attention (with RADIO adapted DINO/SAM) \cite{lan2024proxyclip}. \cref{tab:attention-ablation} highlights how leveraging RADIO can provide significant consistent improvements in mIoU, and efficiency, \textbf{showing RADIO's strong potential as a backbone for OVSS}.

\subsection{Parameter and Compute Efficiency}
\label{sec:experiments_efficiency}
To better contextualize the segmentation accuracy improvements, we perform a thorough analysis of each method’s latency and vision parameter efficiency. For a holistic view that reflects all computations that enabled a method to obtain its mIoU results, we report average latency across all datasets at mid resolution on a V100-32GB GPU at FP32 precision. As illustrated in \cref{fig:fig1}, \methodname{}-base surpasses the strongest baselines, TextRegion Huge and Trident Huge, achieving 9.3-12.6× lower latency while requiring 8.1-12.7× fewer parameters. Furthermore, \methodname{}+, which enhances \methodname{} with SAM-based refinement, delivers more fine-grained segmentation maps while remaining 3.0-4.1× faster and using 7.2-11.3× fewer parameters.

\section{Conclusion}
\label{sec:conclusion}
We present \methodname{}, a dense language-aligned encoder that leverages the RADIO agglomerative framework. \methodname{} effectively employs self correlations for attention and window aggregation, and RADIO SAM adapted features for efficient mask refinement. We evaluate \methodname{} on 8 datasets in 2D and 3D showing significant improvements in zero-shot open-vocabulary semantic segmentation (ZSOVSS) across mIoU, latency, and parameter efficiency. In addition, we provide the first empirical study on the use of RADIO for ZSOVSS highlighting its efficiency and strong emergent dense language alignment. We expect this work to motivate broader exploration of agglomerative models towards efficient, accurate, and generalizable open-vocabulary semantic segmentation. Our future work aims to incorporate instance differentiation, and multi-label segmentation.

{
    \small
    \bibliographystyle{ieeenat_fullname}
    \bibliography{main}
}

\clearpage
\setcounter{page}{1}
\maketitlesupplementary

\setcounter{section}{0}
\setcounter{equation}{0}
\setcounter{figure}{0}
\setcounter{table}{0}

\renewcommand{\thesection}{A\arabic{section}}
\renewcommand{\thesubsection}{A\arabic{subsection}}
\renewcommand{\thefigure}{A.\arabic{figure}}
\renewcommand{\thetable}{A.\arabic{table}}
\renewcommand{\theequation}{A.\arabic{equation}}

\section*{Limitations}
While \methodname{}-base delivers strong mIoU gains with a lightweight vision encoder, it relies on a huge text encoder, unlike CLIP-based baselines in the same model size class. However, text features are computed once per query, so their cost is amortized across many images which. Furthermore, although we demonstrate \methodname{}'s strong performance on eight 2D and 3D OVSS datasets, current OVSS benchmarks remain limited. They primarily test whether each pixel is more similar to query A, B, \textbf{or} C. A more challenging setting would prevent access to all class labels and evaluate queries individually (is this pixel similar to query A?). Moreover, support for multi-label segmentation is not explored. Developing and evaluating on such challenging settings is left for future work.

\section*{Acknowledgments}
\ifnotanon
This work was supported by Defense Science and Technology Agency (DSTA) Contract \#DST000EC124000205, King Abdulaziz University, National Institute on Disability, Independent Living, and Rehabilitation Research (NIDILRR) Grant \#90IFDV0042, and DENSO Corporation Grant \#OSP00016426. The work additionally used Bridges-2 at PSC through allocation cis220039p from the Advanced Cyberinfrastructure Coordination Ecosystem: Services \& Support (ACCESS) program which is supported by NSF grants \#2138259, \#2138286, \#2138307, \#2137603, and \#213296. We also gratefully acknowledge NVIDIA for providing GPUs to Airlab through academic hardware grants. Finally, we thank Mike Ranzinger, Nikhil Keetha, and Parv Maheshwari for insightful discussions.
\fi

\section{Contribution Statement}

\textbf{Omar Alama} Led and shaped the research, conceived the initial ideas for the encoder, wrote the majority of the manuscript, designed key figures and table formats. Additionally, developed the 3D evaluation pipeline and provided coding support and debugging throughout the project.

\noindent \textbf{Darshil Jariwala} Developed the encoder code and refined the initial ideas, conducted all evaluations and ablations for 2D OVSS -- including porting and optimizing 2D baselines, generated qualitative 2D results, and helped with paper writing.

\noindent \textbf{Avigyan Bhattacharya} Conducted all evaluations for 3D OVSS, including porting 3D baselines, generated 3D qualitative figures, adapted ResCLIP to RADIO, wrote the 2D and 3D experimental sections. Contributed to paper writing and refinement.

\noindent \textbf{Seungchan Kim} Incorporated and processed the ScanNet++ dataset for 3D OVSS evaluation, and played a key role in refining the manuscript, improving clarity, structure, and presentation. 

\noindent \textbf{Wenshan Wang} Provided valuable feedback on research design, manuscript writing, and method presentation. 

\noindent \textbf{Sebastian Scherer} Provided valuable feedback on research direction, manuscript clarity, and figure presentation. 

\section{Additional 2D Evaluation Details}
\begin{table}[h]
\caption{Adopted resolution and sliding window settings. Cell format: Shorter side - Crop - Stride. TextRegion's stride always equals the crop size as per their method.}
\centering
\resizebox{\linewidth}{!}{
\begin{tabular}{lccc}
\toprule
\textbf{Dataset} & \textbf{Low \cite{hajimiri2025naclip, wang2024sclip, yang2025resclip}} & \textbf{Mid \cite{shi2024trident}} & \textbf{High \cite{xiao2025textregion}} \\
\midrule
VOC        & 336-224-112 & 336-336-112 & 672-336-336 \\
Stuff   & 336-224-112         & 448-336-224 & 896-336-336 \\
CTX/ADE & 336-224-112        & 576-336-224 & 672-336-336 \\
City   & 560-224-112 & 688-336-224 & 1344-336-336 \\
\bottomrule
\end{tabular}
}

\label{tab:2d_config}
\end{table}

\begin{table*}[!t]
\centering
\caption{2D zero-shot open-vocabulary semantic segmentation mIoU results across resolutions and model sizes. Ranking shown as \colorbox{green!25}{\textbf{first}}, \colorbox{yellow!30}{second}, and \colorbox{orange!30}{third}. One ranking for base and huge is reported to highlight that \methodname{}-base is also superior to huge baselines.}
\resizebox{\linewidth}{!}{
\setlength{\tabcolsep}{3.5pt}
\begin{tabular}{l|cccccc|cccccc|cccccc}
\hline
\multirow{2}{*}{\textbf{Methods}} &
\multicolumn{6}{c|}{\textbf{= Low Resolution}} &
\multicolumn{6}{c|}{\textbf{= Mid Resolution}} &
\multicolumn{6}{c}{\textbf{= High Resolution}} \\
& CTX & VOC & Stuff & ADE & City & Avg
& CTX & VOC & Stuff & ADE & City & Avg
& CTX & VOC & Stuff & ADE & City & Avg \\
\hline
\multicolumn{19}{c}{\textbf{Base Models}} \\
\hline
NACLIP \cite{hajimiri2025naclip}     & 35.17 & 79.71 & 23.3 & 17.42 & 35.49 & 38.22 & 34.7 & 80.29 & 23.64 & 17.78 & 35.98 & 38.48 & 33.86 & 72.57 & 20.04 & 17.8 & 36.74 & 36.20 \\
ResCLIP \cite{yang2025resclip}    & 36.8 & 82.32 & 24.7 & 18.03 & 35.85 & 39.54 & 36.75 & 85.89 & 25.07 & 18.55 & 36.19 & 40.49 & 35.98 & 76.6 & 22.01 & 18.57 & 36.88 & 38.01
 \\
 RayFronts \cite{alama2025rayfronts}  & 36.81 & 72.59 & 25.34 & 23.38 & 36.29 & 38.88 & 38.07 & 80.12 & 27.39 & 24.63 & 40.47 & 42.14 & 36.99 & 67.39 & 23.92 & 24.18 & 39.13 & 38.32 \\
ProxyCLIP \cite{lan2024proxyclip}   & 38.74 & 78.18 & 26.11 & 19.71 & 39.69 & 40.49 & 37.88 & 80.31 & 26.41 & 19.67 & 40.39 & 40.93 & 34.85 & 70.44 & 21.63 & 19.18 & 38.84 & 36.99
 \\
SC-CLIP \cite{bai2024scclip}    & 40.12 & 84.29 & 26.62 & 20.06 & 41.02 & 42.42 & 39.87 & 87.67 & 27.25 & 20.68 & 40.49 & 43.19 & 38.68 & 77.67 & 22.89 & 20.34 & 41.24 & 40.16 \\
Talk2Dino \cite{barsellotti2024talk2dino}  &39.43 & 85.68 & 27.43 & 20.42 & 38.05 & 42.20 & 40.31 & 85.66 & 27.89 & 21.67 & 39.47 & 43.00 & 40.23 & 84.15 & 27.06 & 21.70 & 41.15 & 42.86 \\
Trident  \cite{shi2024trident}   & 41.62 & 83.74 & 28.22 & 21.17 & 41.86 & 43.32 & 42.16 & 84.5 & 28.24 & 21.98 & 42.08 & 43.79 & 40.70 & 80.75 & 25.45 & 20.81 & 42.19 & 41.98 \\
TextRegion \cite{xiao2025textregion}  & 41.53 & 83.83 & 27.39 & 21.21 & 40.49 & 43.17 & 42.29 & 84.21 & 27.13 & 21.74 & 41.26 & 43.33 & 42.88 & 83.19 & 28.62 & 22.55 & 42.15 & 43.88 \\
\methodname{} & \cellcolor{orange!30}{\textbf{44.49}} & \textbf{87.24} & \textbf{29.79} & \cellcolor{orange!30}{\textbf{27.16}} & \textbf{42.04} & \textbf{46.14} & \cellcolor{orange!30}{\textbf{45.64}} & \textbf{89.28} & \cellcolor{yellow!30}{\textbf{30.76}} & \cellcolor{orange!30}{\textbf{28.96}} & \textbf{45.35} & \cellcolor{orange!30}{\textbf{48.00}} & \textbf{45.14} & \textbf{84.07} & \textbf{29.05} & \cellcolor{orange!30}{\textbf{28.93}} & \cellcolor{orange!30}{\textbf{48.79}} & \textbf{47.20} \\
\methodname{}\textbf{+} & \cellcolor{green!25}{\textbf{48.24}} & \textbf{88.46} & \cellcolor{green!25}{\textbf{31.69}} & \cellcolor{green!25}{\textbf{29.63}} & \cellcolor{orange!30}{\textbf{45.58}} & \cellcolor{green!25}{\textbf{48.84}} & \cellcolor{green!25}{\textbf{48.48}} & \cellcolor{yellow!25}{\textbf{90.14}} & \cellcolor{green!25}{\textbf{32.48}} & \cellcolor{green!25}{\textbf{30.83}} & \cellcolor{green!25}{\textbf{48.10}} & \cellcolor{green!25}{\textbf{50.01}} & \cellcolor{yellow!25}{\textbf{47.86}} & \textbf{85.70} & \cellcolor{orange!30}{\textbf{30.43}} & \cellcolor{green!25}{\textbf{30.86}} & \cellcolor{green!25}{\textbf{50.96}} & \cellcolor{yellow!30}{\textbf{49.49}} \\
\hline
\multicolumn{19}{c}{\textbf{Huge Models}} \\
\hline
RayFronts \cite{alama2025rayfronts}  & 31.67 & 72.59 & 23.31 & 20.46 & 28.98 & 35.40 & 32.29 & 73.36 & 23.44 & 21.19 & 31.84 & 36.42 & 32.19 & 67.64 & 22.13 & 21.08 & 34.27 & 35.46 \\
ProxyCLIP \cite{lan2024proxyclip}   & 39.16 & 78.02 & 26.19 & 23.90 & 43.64 & 42.18 & 38.31 & 83.03 & 27.76 & 24.05 & 43.92 & 43.21 & 37.03 & 71.66 & 21.81 & 23.65 & 43.09 & 39.45
 \\
Trident  \cite{shi2024trident}   & 43.16 & 87.97 & 28.55 & 25.64 & \cellcolor{yellow!30}{46.87} & 46.44 & 44.32 & 88.67 & 28.52 & 26.70 & \cellcolor{orange!30}{46.30} & 46.90 & 43.77 & 87.22 & 25.72 & 27.02 & 47.34 & 46.21 \\
TextRegion \cite{xiao2025textregion} & 44.04 & \cellcolor{orange!30}{89.53} & \cellcolor{yellow!30}{30.19} & 24.53 & \cellcolor{green!25}{47.35} & \cellcolor{yellow!30}{47.13} & 44.76 & 89.42 & 29.85 & 26.42 & \cellcolor{yellow!30}{47.04} & 47.50 & \cellcolor{orange!30}{46.13} & \cellcolor{yellow!25}{89.36} & \cellcolor{green!25}{31.22} & 27.30 & 46.88 & \cellcolor{orange!30}{48.18} \\
\methodname{} & \textbf{42.27} & \cellcolor{yellow!30}{\textbf{89.58}} & \textbf{28.3} & \textbf{25.96} & \textbf{38.85} & \textbf{44.99} & \textbf{44.8} & \cellcolor{orange!30}{\textbf{89.74}} & \textbf{28.93} & \textbf{28.21} & \textbf{42.93} & \textbf{46.92} & \textbf{45.01} & \cellcolor{orange!30}{\textbf{88.52}} & \textbf{29.46} & \textbf{27.15} & \textbf{47.75} & \textbf{47.58} \\ 
\methodname{}\textbf{+} & \cellcolor{yellow!30}{\textbf{45.75}} & \cellcolor{green!25}{\textbf{90.39}} & \cellcolor{orange!30}{\textbf{30.17}} & \cellcolor{yellow!30}{\textbf{27.86}} & \textbf{42.49} & \cellcolor{orange!30}{\textbf{47.73}} & \cellcolor{yellow!30}{\textbf{47.76}} & \cellcolor{green!30}{\textbf{90.44}} & \cellcolor{orange!30}{\textbf{30.50}} & \cellcolor{yellow!30}{\textbf{30.09}} & \textbf{46.48} & \cellcolor{yellow!30}{\textbf{49.38}} & \cellcolor{green!30}{\textbf{47.98}} & \cellcolor{green!30}{\textbf{89.40}} & \cellcolor{yellow!30}{\textbf{30.79}} & \cellcolor{yellow!30}{\textbf{29.99}} & \cellcolor{yellow!30}{\textbf{50.58}} & \cellcolor{green!25}{\textbf{50.19}} \\
\hline
\end{tabular}
}
\label{tab:ovss_2d_eq}
\end{table*}

\textbf{Emphasis on resolution standardization}. By examining the settings used for evaluation in previous works, we observe varying resolution standards, not only across methods and datasets, but even within different sub-modules of a method itself. These inconsistencies can conflate performance gains from increased resolutions with method improvements. For example, while NACLIP \cite{hajimiri2025naclip} and ResCLIP \cite{yang2025resclip} choose the low resolution setting in their evaluations, \cref{tab:ovss_2d_lte} shows that they benefit from increased resolutions at the mid setting, making it unfair to compare their low-resolution performance with mid or high resolution performances of other approaches, which is done in many previous works \cite{shi2024trident, xiao2025textregion}. Furthermore, previous works have shown that increasing resolution can hurt performance \cite{shi2024trident, barsellotti2024talk2dino}. Thus, for a fair standard comparison, we rerun all baselines with 3 different representative resolution and sliding window standards (low, mid, high) shown in \cref{tab:2d_config}. Note that evaluation resolution, at which we compute metrics, always remains at the ground truth resolution. To avoid penalizing approaches that perform better when using lower resolutions, \textbf{we report the maximum performance at a resolution limit} in \cref{tab:ovss_2d_lte}. \cref{tab:ovss_2d_lte} answers what the best performance is for a method given a \textbf{resolution limit}, while its source data, shown in \cref{tab:ovss_2d_eq}, answers what the performance is at a \textbf{particular resolution}. While \cref{tab:ovss_2d_lte} is more relevant for method comparisons, we report \cref{tab:ovss_2d_eq} for completeness. Note that \textbf{all other} ablations and tables use mid resolution only.

\section{Additional 3D Evaluation Details}

\textbf{3D mapping for evaluation.} We follow RayFronts' \cite{alama2025rayfronts} approach to construct the 3D map and perform the evaluation. Given a pose $P_t \in SE(3)$, a corresponding depth map $D_t \in \mathbb{R}^{H\times W}$, and a feature map $F_t \in \mathbb{R}^{H\times W \times D}$, feature pixels are back projected to a 3D point cloud $P^{\text{local}}_t = \{(p_i, f_i)\}_{i=1}^M$, where $p_i \in \mathbb{R}^3$ denotes 3D position, and $f_i \in \mathbb{R}^{D+1}$ represents the concatenated feature vector, and hit count (initialized to 1 per point). Local updates are accumulated over frames and voxelized at a resolution of $\alpha$ to form the global semantic voxel map $P^{\text{global}}_t = \{(p_i, f_i)\}_{i=1}^N$. During voxelization, points that lie in the same voxel get their features averaged, and their hit counts summed to use as weights for subsequent averaging. In case of probability space 3D aggregation, the embedding dimension D in the feature map and semantic voxel map is equal to the number of classes as segmentation probabilities are projected instead of embeddings. 

\textbf{Datasets.} Following prior works \cite{gu2024conceptgraphs, alama2025rayfronts, conceptfusion, werby2024hovsg} for our 3D OVSS evaluation, we choose the scenes, \texttt{office[0-4]}, \texttt{room[0-2]} from Replica and \texttt{scene[0011,0050,0231,0378,0518]} from ScanNet. Additionally, to assess performance on a cleaner real-world dataset, we evaluate on ScanNet++ and select nine diverse scenes. The selected scenes are \texttt{scene[00777c41d4, bcd2436daf, a5114ca13d, 2b1dc6d6a5, d551dac194, f9f95681fd, ea42cd27e6, 20ff72df6e, bf6e439e38]}. We use all 101 classes from Replica, the standard ScanNet-to-NYU40 label mapping provided with the dataset itself for ScanNet, and the top 100 classes from ScanNet++ as defined in its official semantic segmentation benchmark. In ScanNet, we assign three of the forty classes (‘otherprop,’ ‘otherstructure,’ and ‘otherfurniture’) as ignore classes due to their ambiguity.

\textbf{Implementation details.} For all the datasets, we resize each image by setting the shorter side to 640 and keeping the original aspect ratio. A frame skip of 10 and 5cm voxels are used for constructing the 3D map. We use external ground-truth voxels for Replica (following \cite{alama2025rayfronts}, we use those provided by HOV-SG) and for ScanNet++. For ScanNet, however, we derive the ground truth by lifting the 2D semantic segmentation annotations into 3D. During evaluation, we perform k-NN matching (k = 5) in accordance with the HOV-SG \cite{werby2024hovsg} protocol, assigning each ground-truth voxel the majority label of its nearest neighbors.

\section{Additional Efficiency Evaluation Details}

\begin{table*}[!t]
\centering
\caption{Compute and parameter efficiency across datasets on a V100 GPU at mid resolution and FP32 precision.
The table reports vision encoder parameters, average mIoU across datasets, and per-dataset latency to capture differences arising from varying input resolutions—both across and within datasets. Illustrated visually in \cref{fig:fig1}. At mid resolution, \textbf{\methodname{}-base surpasses huge Trident and TextRegion baselines while being 9.3x-12.6x faster and having 8.1x-12.7x fewer parameters}.}
\setlength{\tabcolsep}{3.5pt}
\begin{tabular}{l|ccccc|ccc}
\hline
\multirow{2}{*}{\textbf{Methods}} &
\multicolumn{5}{c|}{Latency (s)}  & Lat (s) & Params (M) & mIoU (\%) \\
& CTX & VOC & Stuff & ADE & City & \multicolumn{3}{c}{Avg}\\
\hline

\multicolumn{9}{c}{\textbf{Base Models}} \\
\hline
NACLIP \cite{hajimiri2025naclip} & 0.091 & 0.025 & 0.074 & 0.111 & 0.144 & 0.089 & 86 & 38.48 \\
ResCLIP \cite{yang2025resclip} & 0.252 & 0.152 & 0.266 & 0.325 & 0.311 & 0.261 & 86 & 40.33 \\
RayFronts \cite{alama2025rayfronts} & 0.106 & 0.035 & 0.087 & 0.109 & 0.218 & 0.111 & 103 & 42.14 \\
ProxyCLIP \cite{lan2024proxyclip} & 0.689 & 0.262 & 0.247 & 0.693 & 1.446 & 0.667 & 171 & 40.87 \\
SC-CLIP \cite{bai2024scclip} & 0.148 & 0.059 & 0.111 & 0.174 & 0.243 & 0.147 & 86 & 41.86 \\
Trident \cite{shi2024trident} & 0.353 & 0.229 & 0.348 & 0.414 & 0.923 & 0.454 & 264 & 43.79 \\
TextRegion \cite{xiao2025textregion} & 0.410 & 0.403 & 0.438 & 0.428 & 1.729 & 0.682 & 167 & 43.33 \\
Talk2Dino \cite{barsellotti2024talk2dino} & 0.150 & 0.043 & 0.155 & 0.188 & 0.289 & 0.165 & 86 & 43.00 \\
\methodname{} & \textbf{0.109} & \textbf{0.036 }& \textbf{0.088} & \textbf{0.112} & \textbf{0.228} & \textbf{0.115} & \textbf{106} & \cellcolor{orange!30} \textbf{48.00} \\
\methodname{}\textbf{+} & \textbf{0.261} & \textbf{0.156} & \textbf{0.286} & \textbf{0.360} & \textbf{0.709} & \textbf{0.354} & \textbf{119 }& \cellcolor{green!25} \textbf{50.01} \\
\hline
\multicolumn{9}{c}{\textbf{Huge Models}} \\
\hline
RayFronts \cite{alama2025rayfronts} & 0.619 & 0.204 & 0.460 & 0.598 & 1.362 & 0.649 & 661 & 36.40 \\
ProxyCLIP \cite{lan2024proxyclip} & 1.297 & 0.494 & 0.451 & 1.28 & 3.168 & 1.338 & 717 & 43.21 \\
Trident \cite{shi2024trident}  & 1.426 & 0.900 & 1.123 & 1.444 & 2.365 & 1.451 & 1349 & 46.90 \\
TextRegion \cite{xiao2025textregion} & 0.732 & 0.670 & 0.711 & 0.744 & 2.476 & 1.067 & 857 & 47.50 \\
\methodname{} & \textbf{0.617} & \textbf{0.202} & \textbf{0.457} & \textbf{0.594} & \textbf{1.363} & \textbf{0.647} & \textbf{664} & \textbf{46.92} \\
\methodname{}\textbf{+} & \textbf{1.172} & \textbf{0.734} & \textbf{1.074} & \textbf{1.341} & \textbf{2.128} & \textbf{1.290} & \textbf{677} & \cellcolor{yellow!30} \textbf{49.38} \\
\hline
\end{tabular}
\label{tab:efficiency}
\end{table*}

\textbf{Improving baselines efficiency.} To have a robust comparison, we modify the inference code of NACLIP \cite{hajimiri2025naclip}, ResCLIP \cite{yang2025resclip}, SC-CLIP \cite{bai2024scclip}, and ProxyCLIP \cite{lan2024proxyclip} to use batched sliding window inference as opposed to iterating over each window. This significantly improves the baselines latency and gives us stronger comparison points. In our experiments, all baselines use batched sliding windows with a batch size equal to the number of windows.

\textbf{What latency to report?} Input resolutions vary across datasets and even across samples since only the shorter image side is fixed. Some methods adjust hyperparameters per dataset, affecting total computation. Furthermore, the latency of methods (including \methodname{}+) can vary with the content of an image as the number of masks varies. To provide a holistic measure that reflects all computations contributing to a method’s final segmentation accuracy, we report the average latency across all validation samples in each dataset, as shown in \cref{tab:efficiency}. The table reveals substantial latency differences between datasets, underscoring that measuring latency on a single image or at a fixed resolution does not capture overall performance. 

Latencies are measured using mid resolution, FP32 precision, on a Tesla V100-32GB GPU. In addition, the number of parameters of the vision encoders of each method is reported. \cref{fig:fig1} visualizes the mIoU vs number of parameters and mIoU vs latency tradeoffs. Notably, at mid resolution, \textbf{\methodname{}-base surpasses huge Trident and TextRegion baselines while being 9.3x-12.6x faster and having 8.1x-12.7x fewer parameters}.

\section{Additional Ablations and Detailed Tables}

\begin{table}[h]
\caption{\textbf{SCRA+SCGA generalize} consistently across all RADIO versions and sizes improving mIoU. Rv3-l is excluded due to lack of CLS-patch alignment.}
\label{tab:generalization}
\centering
\resizebox{\linewidth}{!}{
\begin{tabular}{ll cccccc}
\toprule
Version & Method & CTX & VOC & Stuff & ADE & City & Avg \\
\midrule
\multirow{3}{*}{Rv2.5-b} & Base & 34.24 & 84.99 & 24.66 & 23.11 & 32.78 & 39.96 \\
 & {\scriptsize +SCRA+SCGA} & 44.52 & 87.59 & 30.52 & 28.27 & 43.56 & 46.89 \\
 & $\Delta$ & \textbf{+10.28} & \textbf{+2.60} & \textbf{+5.86} & \textbf{+5.16} & \textbf{+10.78} & \textbf{+6.93} \\
\midrule
\multirow{3}{*}{Rv2.5-l} & Base & 31.70 & 83.88 & 23.68 & 21.00 & 32.49 & 38.55 \\
 & {\scriptsize +SCRA+SCGA} & 38.85 & 89.02 & 28.14 & 25.53 & 43.72 & 45.05 \\
 & $\Delta$ & \textbf{+7.15} & \textbf{+5.14} & \textbf{+4.46} & \textbf{+4.53} & \textbf{+11.23} & \textbf{+6.50} \\
\midrule
\multirow{3}{*}{Rv2.5-h} & Base & 31.12 & 79.04 & 23.00 & 21.92 & 29.78 & 36.97 \\
 & {\scriptsize +SCRA+SCGA} & 38.67 & 82.52 & 27.31 & 25.81 & 36.63 & 42.19 \\
 & $\Delta$ & \textbf{+7.55} & \textbf{+3.48} & \textbf{+4.31} & \textbf{+3.89} & \textbf{+6.85} & \textbf{+5.22} \\
\midrule
\multirow{3}{*}{Rv3-b} & Base & 40.40 & 88.07 & 27.41 & 27.30 & 39.94 & 44.62 \\
 & {\scriptsize +SCRA+SCGA} & 45.64 & 89.28 & 30.76 & 28.96 & 45.35 & 48.00 \\
 & $\Delta$ & \textbf{+5.24} & \textbf{+1.21} & \textbf{+3.35} & \textbf{+1.66} & \textbf{+5.41} & \textbf{+3.38} \\
\midrule
\multirow{3}{*}{Rv3-h} & Base & 35.23 & 88.35 & 24.55 & 24.34 & 33.74 & 41.24 \\
 & {\scriptsize +SCRA+SCGA} & 44.80 & 89.74 & 28.93 & 28.21 & 42.93 & 46.92 \\
 & $\Delta$ & \textbf{+9.57} & \textbf{+1.39} & \textbf{+4.38} & \textbf{+3.87} & \textbf{+9.19} & \textbf{+5.68} \\
\bottomrule
\end{tabular}}
\end{table}

\textbf{RADSeg modules generalize across RADIO}. \cref{tab:generalization} demonstrates that the proposed SCRA and SCGA modules generalize across all RADIO versions and model sizes, with consistent improvements ranging from +3.4 to +6.9 avg mIoU. Gains are largest on Cityscapes (up to +11.2), whose high-resolution images require more sliding windows and thus benefit most from SCGA's cross-window consistency. V2.5 variants also see larger gains overall, where weaker base spatial locality leaves more room for refinement.

\begin{figure}[tbp]
\includegraphics[width=\linewidth]{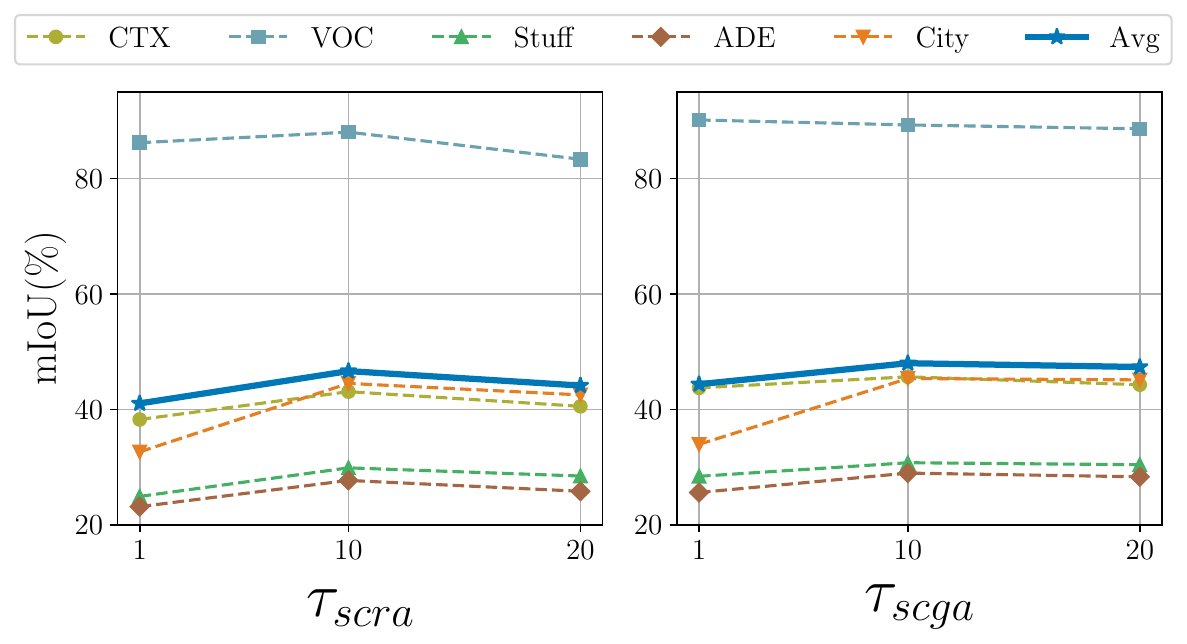}
\caption{Ablation study on different temperature parameters $\tau_{scra}$ and $\tau_{scga}$. Left plot shows performance as we change $\tau_{scra}$ without SCGA. Right plot uses $\tau_{scra}=10$ and varies $\tau_{scga}$. Overall $\tau_{scra}=\tau_{scga}=10$ yield the best results on average.}
\label{fig:tau_ablations}
\end{figure}

\textbf{Scaling the attention correlation matrix has a notable impact}. We ablate temperature parameters $\tau_{scra}$ and $\tau_{scga}$ for SCRA and SCGA at mid resolution across 2D datasets. \cref{fig:tau_ablations} highlights the importance of scaling the attention correlation matrix to sharpen the attention to semantically similar patches and similarly for global aggregation. $\tau_{scra} = \tau_{scga} = 10$ yield the best results and is what we use for all experiments.

\begin{table}[h]
\caption{Increasing SCRA recursion does not help (w/SCGA).}
\label{tab:recursion}
\centering
\begin{tabular}{lccc}
\toprule
Iterations & 1 & 2 & 3 \\
\midrule
Avg mIoU & \textbf{48.0} & 47.5 & 47.0 \\
\bottomrule
\end{tabular}
\end{table}
\textbf{Increasing recursion depth of SCRA does not help}. As shown in \cref{tab:recursion}, increasing recursion beyond a single iteration yields diminishing returns, with avg mIoU dropping from 48.0 to 47.0 at 3 iterations, likely due to over-reliance on the correlation signal.

\begin{table}[h]
\caption{RADSeg supports single-pass high-res inference thanks to RADIO's flexible resolution support, unlike CLIP-based methods that require sliding windows. (mIoU shown)}
\label{tab:sliding}
\centering
\resizebox{\linewidth}{!}{
\setlength{\tabcolsep}{3pt}
\begin{tabular}{l|ccccc|c}
\hline
\textbf{Method} & CTX & VOC & Stuff & ADE & City & Avg \\
\hline
Sliding Window & \cellcolor{orange!30}45.64 & \cellcolor{green!25}\textbf{89.28} & \cellcolor{orange!30}30.76 & \cellcolor{yellow!30}28.96 & \cellcolor{green!25}\textbf{45.35} & \cellcolor{green!25}\textbf{48.00} \\
Single Inference & \cellcolor{green!25}\textbf{47.52} & \cellcolor{yellow!30}89.02 & \cellcolor{green!25}\textbf{31.47} & \cellcolor{green!25}\textbf{29.51} & \cellcolor{orange!30}41.96 & \cellcolor{yellow!30}47.90 \\
Single Infer w/o SCGA & \cellcolor{green!25}\textbf{47.52} & \cellcolor{orange!30}87.81 & \cellcolor{yellow!30}31.29 & \cellcolor{orange!30}28.81 & \cellcolor{yellow!30}43.83 & \cellcolor{orange!30}47.85 \\
\hline
\end{tabular}}
\end{table}

\textbf{RADSeg supports single-pass high-resolution inference}. As shown in \cref{tab:sliding}, single-pass inference only entails a $-$0.1 avg mIoU drop. Unlike CLIP-based baselines, whose fixed positional embeddings necessitate sliding windows at higher resolutions, RADIO's cropped position embeddings (CPE) natively accept arbitrary resolutions, making single-pass inference practical. Notably, single inference outperforms sliding windows on most datasets, with only Cityscapes—the highest resolution dataset requiring the most windows—benefiting from the sliding window approach. \textbf{SCGA remains beneficial even without windows by refining global feature consistency}.

\begin{table*}[!htbp]
\centering
\caption{Expanded version of \cref{tab:backbone_adaptor_ablations}. Ablation of different RADIO language adapters across
model sizes. Mid resolution is used. \textbf{RADIOv3-base with the SigLIP2 CLS adaptor has the best performance across datasets and model sizes}.}
\resizebox{\linewidth}{!}{
\setlength{\tabcolsep}{3pt}
\begin{tabular}{l|cccccc|cccccc|cccccc}
\hline
\multirow{2}{*}{\textbf{Methods}} & \multicolumn{6}{c|}{Base} & \multicolumn{6}{c|}{Large} & \multicolumn{6}{c}{Huge} \\
 & CTX & VOC & Stuff & ADE & City & Avg & CTX & VOC & Stuff & ADE & City & Avg & CTX & VOC & Stuff & ADE & City & Avg \\
\hline
Rv2.5-SigLIP$_{\text{cls}}$ & \cellcolor{orange!30} 34.24 & 84.99 & \cellcolor{orange!30} 24.66 & \cellcolor{orange!30} 23.11 & \cellcolor{orange!30} 32.78 & \cellcolor{orange!30} 39.96 & \cellcolor{green!25} 31.70 & \cellcolor{yellow!30} 83.88 & \cellcolor{green!25} 23.68 & \cellcolor{green!25} 21.00 & \cellcolor{green!25} 32.49 & \cellcolor{green!25} 38.55 & \cellcolor{yellow!30} 31.12 & 79.04 & \cellcolor{yellow!30} 23.00 & \cellcolor{yellow!30} 21.92 & \cellcolor{yellow!30} 29.78 & \cellcolor{yellow!30} 36.97 \\Rv2.5-CLIP$_{\text{cls}}$ & 33.42 & \cellcolor{orange!30} 86.55 & 23.56 & 22.37 & 31.24 & 39.43 & \cellcolor{yellow!30} 31.09 & \cellcolor{green!25} 84.57 & \cellcolor{yellow!30} 22.68 & \cellcolor{yellow!30} 20.10 & \cellcolor{yellow!30} 30.53 & \cellcolor{yellow!30} 37.79 & 27.81 & \cellcolor{orange!30} 79.70 & 21.08 & 20.03 & 25.65 & 34.85 \\
Rv2.5-SigLIP$_{\text{patch}}$ & 0.31 & 1.00 & 0.06 & 0.14 & 0.09 & 0.32 & 0.27 & 0.97 & 0.06 & 0.12 & 0.14 & 0.31 & 0.37 & 1.06 & 0.06 & 0.17 & 0.13 & 0.36 \\
\textbf{Rv3-SigLIP$_{\text{cls}}$} & \cellcolor{green!25} \textbf{40.40} & \cellcolor{green!25} \textbf{88.07} & \cellcolor{green!25} \textbf{27.41} & \cellcolor{green!25} \textbf{27.30} & \cellcolor{green!25} \textbf{39.94} & \cellcolor{green!25} \textbf{44.62} & \cellcolor{orange!30} \textbf{1.00} & \textbf{1.46} & \cellcolor{orange!30} \textbf{0.52} & \cellcolor{orange!30} \textbf{0.15} & \cellcolor{orange!30} \textbf{1.18} & \cellcolor{orange!30} \textbf{0.86} & \cellcolor{green!25} \textbf{35.23} & \cellcolor{green!25} \textbf{88.35} & \cellcolor{green!25} \textbf{24.55} & \cellcolor{green!25} \textbf{24.34} & \cellcolor{green!25} \textbf{33.74} & \cellcolor{green!25} \textbf{41.24} \\
Rv3-CLIP$_{\text{cls}}$ & \cellcolor{yellow!30} 37.90 & \cellcolor{yellow!30} 86.89 & \cellcolor{yellow!30} 26.36 & \cellcolor{yellow!30} 24.48 & \cellcolor{yellow!30} 35.99 & \cellcolor{yellow!30} 42.32 & 0.67 & 1.71 & 0.07 & 0.12 & 1.17 & 0.75 & \cellcolor{orange!30} 29.48 & \cellcolor{yellow!30} 84.07 & \cellcolor{orange!30} 21.85 & \cellcolor{orange!30} 21.39 & \cellcolor{orange!30} 27.16 & \cellcolor{orange!30} 36.79 \\
Rv3-SigLIP$_{\text{patch}}$ & 0.11 & 2.47 & 0.05 & 0.14 & 0.89 & 0.73 & 0.12 & \cellcolor{orange!30} 2.50 & 0.04 & 0.14 & 0.98 & 0.76 & 0.12 & 0.84 & 0.05 & 0.12 & 0.14 & 0.25 \\
\hline
\end{tabular}}
\label{tab:expanded_backbone_adaptor_ablations}
\end{table*}

\begin{table}[!t]
\centering
\caption{Expanded version of \cref{tab:attention-ablation}. Ablation study showing RADIOv3’s ability to empower existing approaches. Mid resolution is used. ProxyRADIO-D/S refer to using DINOv2 and SAM adapted feature maps respectively for the proxy attention. \textbf{RADIO can improve mIoU for different CLIP-based baselines}.}
\resizebox{\linewidth}{!}{
\setlength{\tabcolsep}{3pt}
\begin{tabular}{l|ccccc|c}
\hline
\textbf{Methods} &CTX &VOC &Stuff &ADE &City &Avg \\
\hline
NACLIP \cite{hajimiri2025naclip} &34.70 &80.28 &23.65 &17.78 &35.98 &38.48 \\
NARADIO &40.85 &86.25 &27.93 & \cellcolor{yellow!30}27.19 &40.47 &44.54 \\
ProxyCLIP \cite{lan2024proxyclip} &37.76 &80.24 &26.4 &19.65 &40.28 &40.87 \\
ProxyRADIO-D &41.62 &86.52 & \cellcolor{orange!30}29.28 &26.90 & \cellcolor{orange!30}42.87 & \cellcolor{orange!30}45.44 \\
ProxyRADIO-S & \cellcolor{yellow!30}41.99 & \cellcolor{orange!30}87.90 & \cellcolor{yellow!30}29.33 & \cellcolor{orange!30}27.00 & \cellcolor{yellow!30}44.04 & \cellcolor{yellow!30}46.05 \\
ResCLIP \cite{yang2025resclip} &36.52 &85.79 &24.91 &18.43 &36.02 &40.33 \\
ResRADIO & \cellcolor{orange!30}41.79 & \cellcolor{yellow!30}87.96 &28.74 &26.63 &40.77 &45.18 \\
\methodname{} & \cellcolor{green!25}\textbf{45.64} & \cellcolor{green!25}\textbf{89.28} & \cellcolor{green!25}\textbf{30.76} & \cellcolor{green!25}\textbf{28.96} & \cellcolor{green!25}\textbf{45.35} & \cellcolor{green!25}\textbf{48.00} \\
\hline
\end{tabular}
}
\label{tab:expanded_attention_ablation}
\end{table}

Finally, for reference, we provide per-dataset details for \cref{tab:backbone_adaptor_ablations} in \cref{tab:expanded_backbone_adaptor_ablations} and per-dataset details for \cref{tab:attention-ablation} in \cref{tab:expanded_attention_ablation}.

\section{CLS-Patch Alignment Analysis}

\label{sec:cls_patch_alignment}

\begin{figure}[t]
    \centering
    \resizebox{0.95\linewidth}{!}{\includegraphics{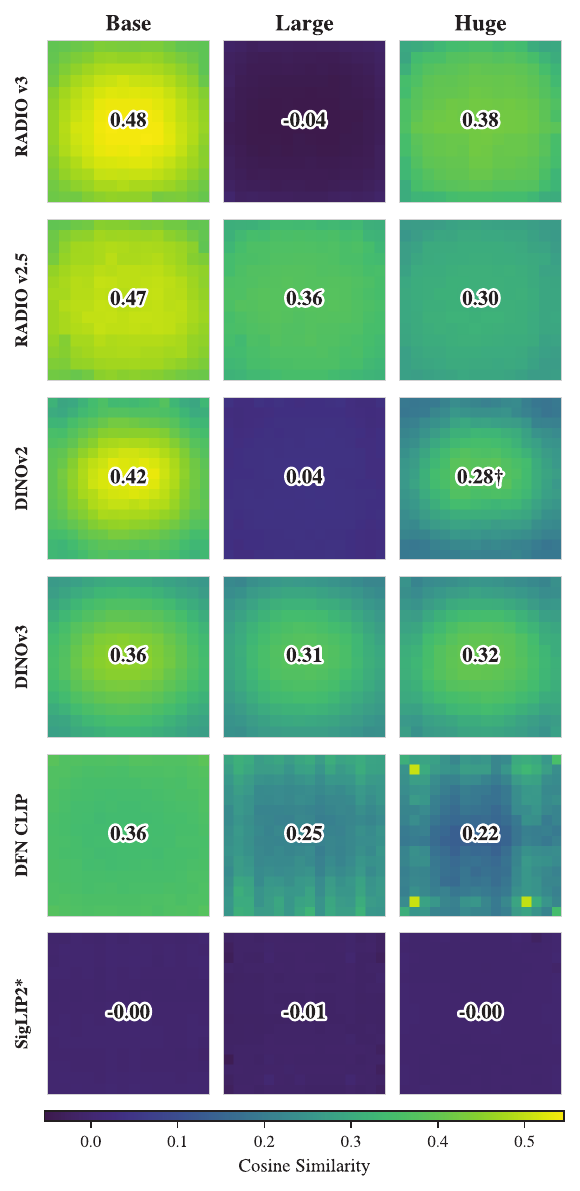}}
    \caption{Average CLS-to-patch cosine similarity across six vision model families at three scales. Lower-capacity models consistently exhibit stronger CLS-to-patch alignment. RADIOv3-L and DINOv2-L are notable outliers with near-zero alignment. SigLIP2 shows no alignment due to its MAP pooling architecture. \textdagger Giant variant (1.1B params). *Pooled output vs.\ patch features.}
    \label{fig:cls_patch_alignment}
\end{figure}

A key property exploited by RADSeg is the alignment between the CLS summary token and output patch features in the RADIO backbone. When this alignment is strong, projecting patch features through the language adaptor head---which is trained to map the CLS token to the text embedding space---yields spatially dense, language-aligned features suitable for zero-shot segmentation. When this alignment is weak, the adaptor projection produces representations that do not correspond to meaningful text embeddings, and segmentation performance degrades.

To quantify this, we compute the average cosine similarity between each output patch feature and the CLS token across 1{,}000 ImageNet validation images for six model families at three scales (\cref{fig:cls_patch_alignment}): RADIOv3 and RADIOv2.5 (using their respective language adaptors), DINOv2 with registers, DINOv3, DFN CLIP, and SigLIP2. 

Two findings emerge. First, RADIOv3-L exhibits near-zero CLS-patch similarity ($-0.04$), indicating a lack of alignment between the CLS summary and spatial patch representations. This directly explains the zero segmentation performance observed for this variant in \cref{tab:backbone_adaptor_ablations}. A similar phenomenon appears in DINOv2-L ($0.04$); understanding why the Large scale specifically loses alignment in both families may require reproducing/retraining these models and hence is beyond our computational resources and scope.

Second, across model families, lower-capacity backbones consistently exhibit stronger CLS-patch alignment (e.g., C-RADIO v3: $0.48$ Base vs.\ $0.38$ Huge; RADIO v2.5: $0.47$ vs.\ $0.30$; DFN CLIP: $0.36$ vs.\ $0.22$). We hypothesize that models with limited capacity are constrained to share representational structure between the CLS and patch tokens, naturally producing stronger alignment. This is consistent with observations in single-encoder vision models~\cite{barsellotti2024talk2dino} and explains the counterintuitive, yet useful, result that base-scale RADIO backbones can outperform larger variants for zero-shot segmentation. 

Note that the spatial pattern of alignment also exhibits a consistent center bias across models, reflecting the tendency of the CLS token to summarize the central, salient entity in the image. Moreover, SigLIP2 shows near-zero similarity (${\sim}0.00$) across all scales, as its MAP pooling head applies learned cross-attention projections that place the pooled output in a different subspace from the output patch features.

\section{Additional Qualitative Results}

We provide additional 2D and 3D qualitative comparisons in \cref{fig:2d_qualitative_extra} and \cref{fig:3d_qualitative_extra}, further illustrating the strengths of \methodname{} over existing open-vocabulary segmentation methods. In 2D, \methodname{} yields cleaner boundaries and more accurate segmentations in both cluttered indoor scenes and complex urban layouts. In 3D, it shows strong multi-view consistency, producing coherent semantic voxels with far fewer outliers and mislabeled regions than the baselines. These visualizations reinforce the ability of \methodname{}, with its proposed components, to suppress noise and enhance object localizations across various scenarios.

\begin{figure*}[!htbp]
\centering
\includegraphics[width=0.8\textwidth]{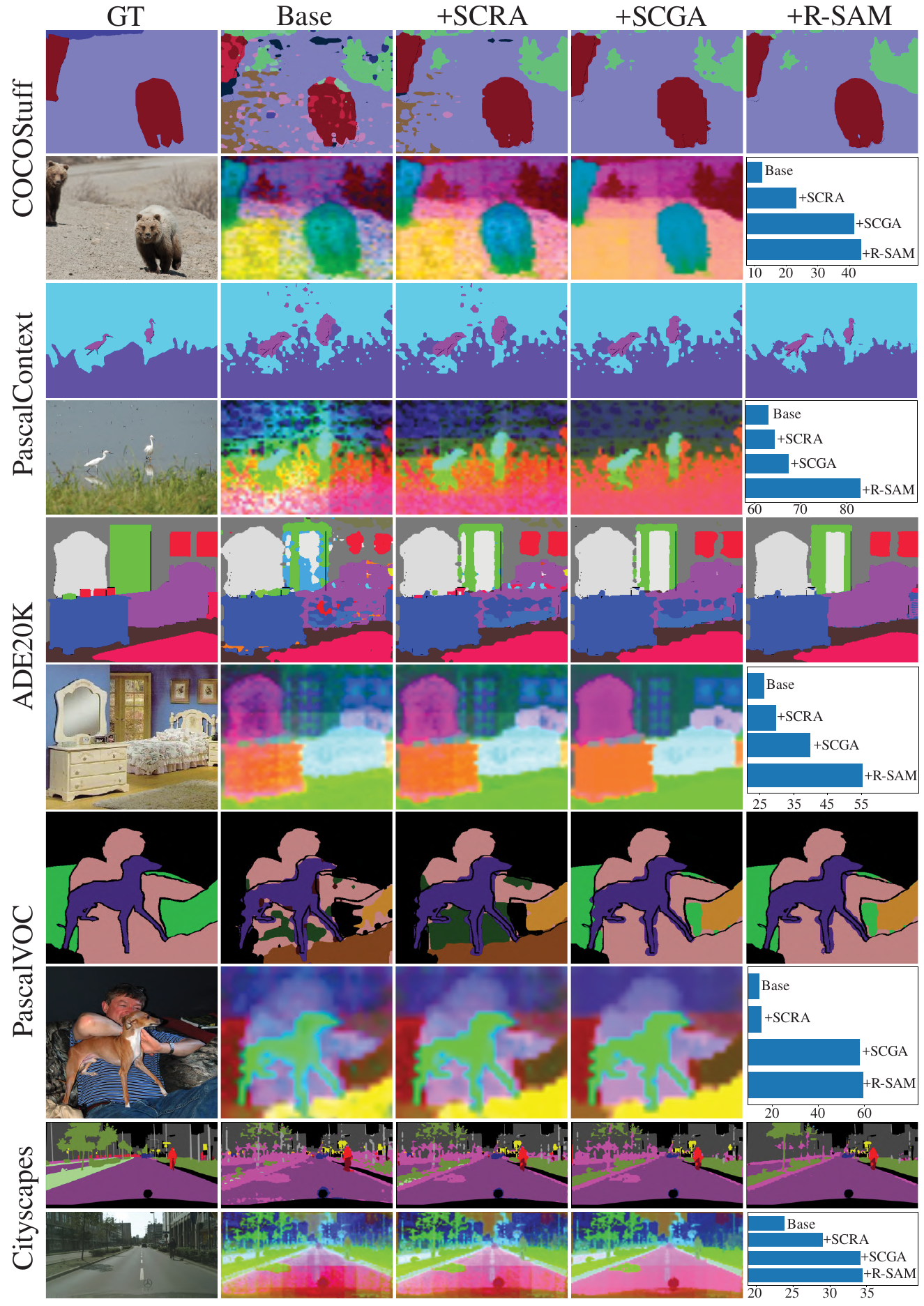}
\caption{Qualitative comparison of the contribution of each \methodname{} component to feature and prediction quality.
For each 2D dataset we show two rows: the first demonstrates how adding \methodname{} components progressively improves segmentation output, while the second (Showing first 3 PCA components) illustrates how SCRA and SCGA enhance feature map quality and mitigate windowing artifacts. The accompanying bar plot links these visual trends to per-sample mIoU scores. Overall, the proposed \methodname{} components yield clear improvements in both segmentation accuracy and feature-map fidelity.}
\label{fig:2d_method_qualitative_ablation}
\end{figure*}

\textbf{RADSeg modules qualitatively improve feature map and segmentation quality}. To demonstrate the effectiveness of the components of \methodname{}, we augment \cref{tab:ablation_hier} with a qualitative visualization. \cref{fig:2d_method_qualitative_ablation} qualitatively illustrates the effectiveness of SCRA and SCGA in suppressing noise in the feature map and segmentation as well as reducing windowing artifacts. The effect is particularly pronounced in higher resolution datasets like Cityscapes. While RADIO-SAM refinement is a post-segmentation process that cannot refine features, it is able to provide higher fidelity masks in many cases. Overall, the proposed \methodname{} and \methodname{}+ components demonstrate quantitative and qualitative improvements.

\begin{figure*}[!htbp]
\centering
\includegraphics[width=\textwidth]{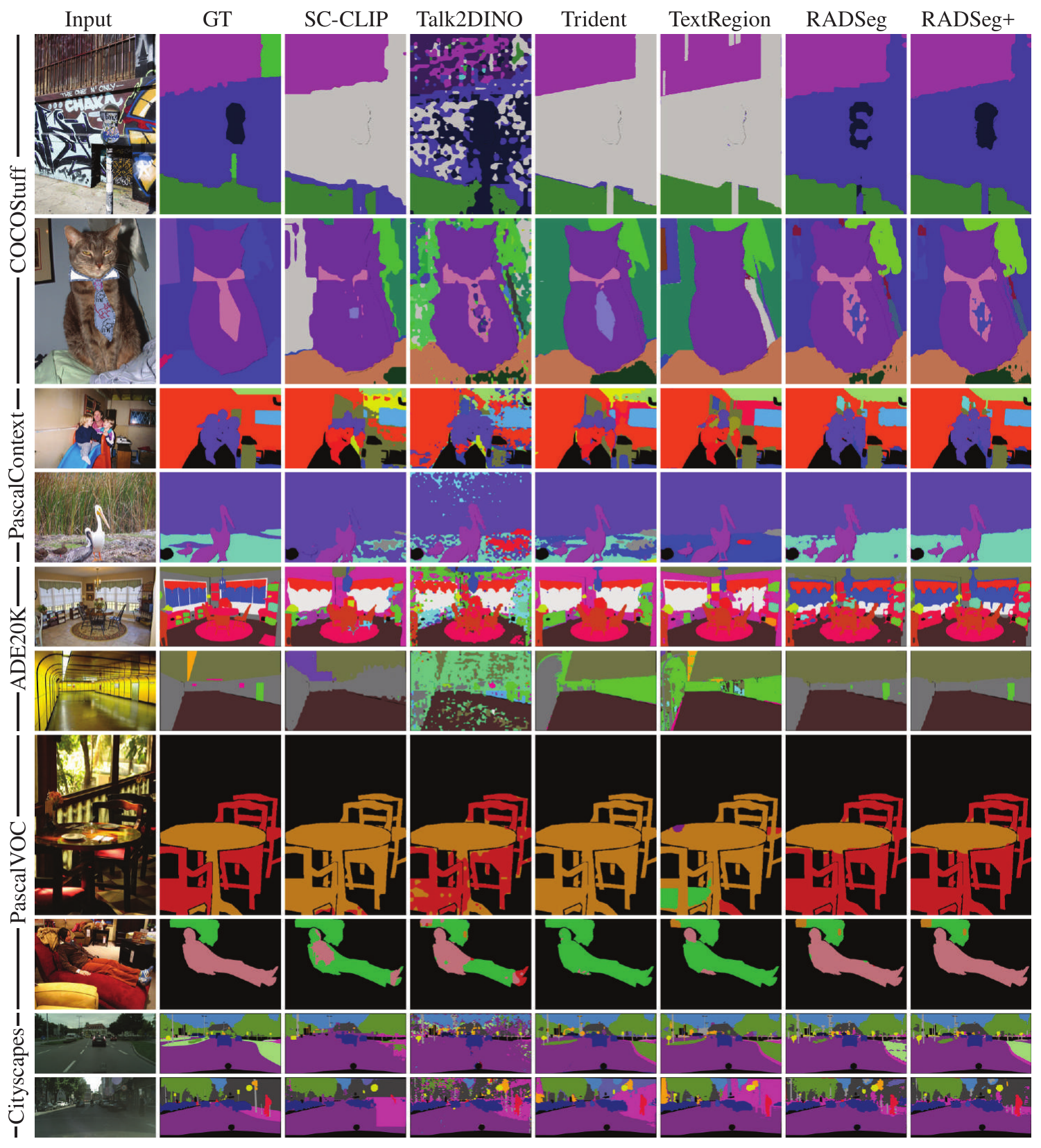}
\caption{Additional visualizations of 2D semantic segmentation results generated by \methodname{}, \methodname{}\textbf{+} and the most competitive baselines on images from all the benchmarks. Along with achieving SOTA mIoU for 2D OVSS, \methodname{} and \methodname{}\textbf{+} produce more precise segmentation maps and sharper object boundaries for both single-object as well as multi-object complex scenes.}
\label{fig:2d_qualitative_extra}
\end{figure*}

\begin{figure*}[!htbp]
\centering
\includegraphics[width=0.7\textwidth]{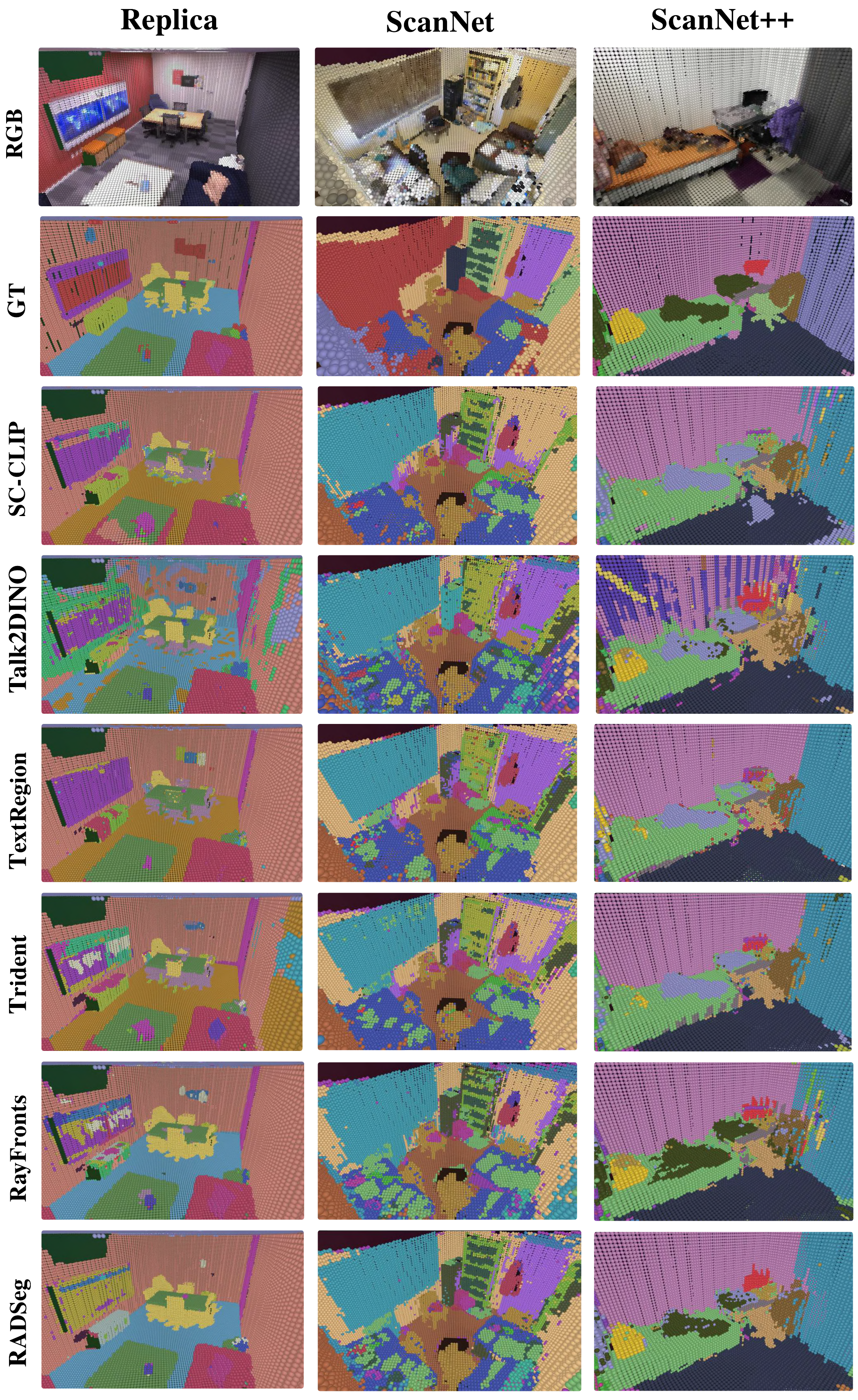}
\caption{Sample visualizations of 3D semantic segmentation results generated by \methodname{} and all the baselines for scenes from Replica (scene-\texttt{office2}), ScanNet (scene-\texttt{0378}), and ScanNet++ (scene-\texttt{ea42cd27e6}. “RGB” and “GT” refer to the RGB scene reconstruction and Ground Truth semantics for each corresponding scene. \methodname{} achieves SOTA mIoU for 3D OVSS and produces cleaner and more accurate semantic voxels.}
\label{fig:3d_qualitative_extra}
\end{figure*}

\end{document}